\pdfoutput=1
\documentclass[11pt]{article}

\usepackage[final]{acl}

\usepackage{times}
\usepackage{latexsym}
\usepackage[T1]{fontenc}
\usepackage[utf8]{inputenc}
\usepackage{microtype}
\usepackage{inconsolata}
\usepackage{natbib}
\usepackage{graphicx}
\usepackage{algorithm}
\usepackage{algorithmic}
\usepackage{amsmath,amsfonts}
\usepackage{array}
\usepackage{microtype}
\usepackage{booktabs}
\usepackage{multirow}
\usepackage{multicol}
\usepackage{makecell}
\usepackage{enumerate}
\usepackage{enumitem}
\usepackage{xcolor}
\usepackage{xspace}
\usepackage{balance}
\usepackage{amsthm}
\usepackage{hyperref}
\usepackage{subcaption}
\usepackage{tcolorbox}
\usepackage{tikz}

\newcommand{\model}{SiPeR\xspace}

\title{\textit{Where and What}: Reasoning Dynamic and Implicit Preferences in \\ Situated Conversational Recommendation}

\author{ 
    Dongding Lin$^{1}$, ~~Jian Wang$^{1,2 \dagger}$, ~~Yongqi Li$^{1}$, ~~Wenjie Li$^{1}$ \\
    $^1$ Department of Computing, The Hong Kong Polytechnic University \\
    $^2$ College of Computer Science, Sichuan University \\
    \texttt{dongding88.lin@connect.polyu.hk} ~~~
    \texttt{jian51.wang@polyu.edu.hk} \\
    \texttt{liyongqi0@gmail.com} ~~~ 
    \texttt{cswjli@comp.polyu.edu.hk}
}

\begin{document}
\maketitle

\renewcommand{\thefootnote}{$\dagger$}
\footnotetext[1]{Corresponding author. This work was mainly conducted at PolyU, while the author is now at Sichuan University.}
\setcounter{footnote}{0}
\renewcommand{\thefootnote}{\arabic{footnote}}

\begin{abstract}
Situated conversational recommendation (SCR), which utilizes visual scenes grounded in specific environments and natural language dialogue to deliver contextually appropriate recommendations, has emerged as a promising research direction due to its close alignment with real-world scenarios.
Compared to traditional recommendations, SCR requires a deeper understanding of dynamic and implicit user preferences, as the surrounding scene often influences users' underlying interests, while both may evolve across conversations. This complexity significantly impacts the timing and relevance of recommendations.
To address this, we propose situated preference reasoning (\model), a novel framework that integrates two core mechanisms: 
(\romannumeral1) \textit{Scene transition estimation}, which estimates whether the current scene satisfies user needs, and guides the user toward a more suitable scene when necessary; and (\romannumeral2) \textit{Bayesian inverse inference}, which leverages the likelihood of multimodal large language models (MLLMs) to predict user preferences about candidate items within the scene. 
Extensive experiments on two representative benchmarks demonstrate \model's superiority in both recommendation accuracy and response generation quality.
The code and data are available at \url{https://github.com/DongdingLin/SiPeR}.

\end{abstract}

\section{Introduction}

Conversational recommendation \citep{li2018towards,gao2021advances,jannach2021survey,zhou2022c2}, as an extensively explored research area, focuses on delivering high-quality recommendations through natural language dialogue. It enables recommenders to actively inquire about user preferences and respond dynamically to user requests. 
In many real-world scenarios, recommendations are inherently grounded in specific environments, such as live promotions in clothing or furniture stores~\citep{kottur2023overview}. 
This has recently shifted research interest to situated conversational recommendation (SCR)~\citep{DBLP:conf/mm/LinWLL24,DBLP:journals/corr/abs-2412-18416}, which leverages visual scenes grounded in specific environments and natural language dialogue to deliver contextually appropriate recommendations. 
This close alignment with the real world underscores the importance of SCR as a promising and practical research direction.

\begin{figure}[t!]
    \centering
    \includegraphics[width=1\linewidth]{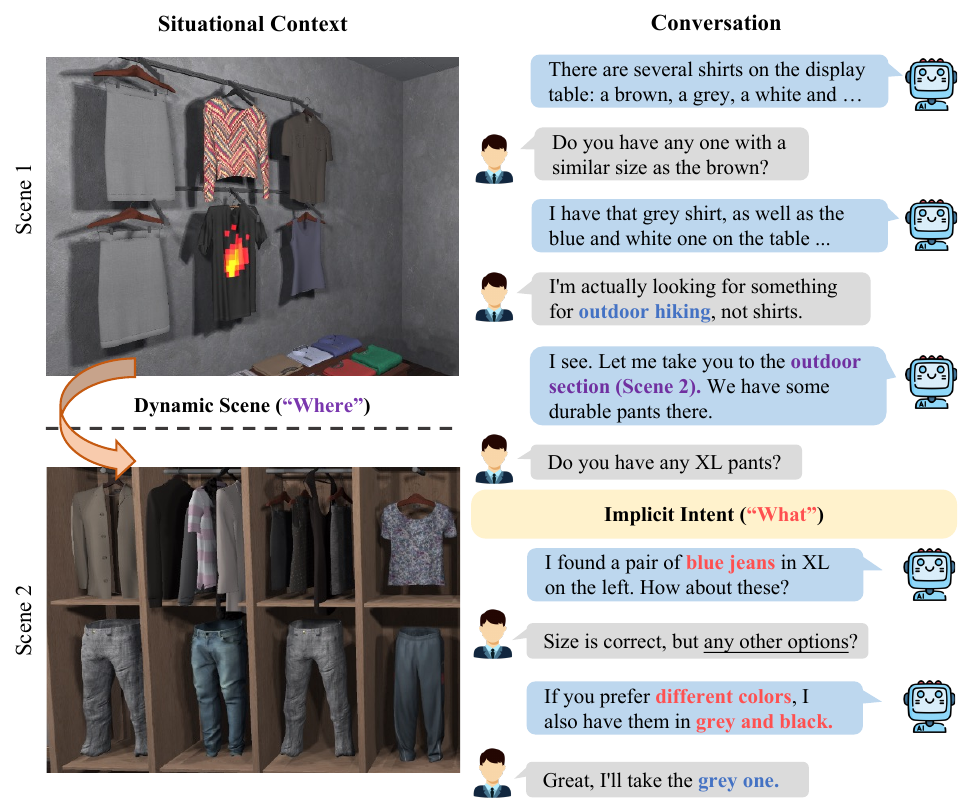}
    \caption{An illustrative example from the repurposed SIMMC 2.1~\citep{kottur2023overview} dataset for situated conversational recommendation, where the interaction process between the user and the virtual assistant is grounded in evolving scenes. In the bottom panel, although the initially suggested blue jeans satisfy the size constraint, the user's final acceptance shows that the grey pants are the ground-truth target, highlighting the need to reason about implicit preferences.}
    \label{fig:example_intro}
    \vspace{-8pt}
\end{figure}

Despite its great potential, existing studies in SCR primarily focused on dataset curation~\citep{DBLP:conf/coling/MoonKCDPLWDBCSG20,kottur2023overview,DBLP:conf/mm/LinWLL24,DBLP:journals/corr/abs-2412-18416}, yet they failed to establish a clear framework for effectively solving the task. 
Building an effective SCR system is non-trivial due to its challenges in reasoning user preferences: 
1) User preferences are often \textbf{dynamic} and varied by situations. In SCR, user interests are often influenced by the surrounding environment.
When the surrounding scene evolves across conversations, user preferences can shift, adding further complexity to the recommendation.
For example, 
as illustrated in the top panel of Figure~\ref{fig:example_intro} (The ``Where''), when the user expresses an interest in ``outdoor hiking,'' the system recognizes that the current formal wear scene is a mismatch. Consequently, it must actively guide the user to the outdoor section (Scene-2) to align with potential user interests.
This necessitates a critical decision-making capability for the system to determine \textit{where} to transition between scenes, which has been largely overlooked in prior work.
2) User preferences are often \textbf{implicit} rather than explicitly stated. 
For instance, in the bottom panel (The ``What''), the user acknowledges that ``the size is correct'' but still asks for other options.
In the full dialogue, the user eventually accepts the grey pants, indicating that while the size constraint is satisfied, the initially recommended blue jeans do not match the user's intended purchase item. To address this, the system must accurately distinguish and predict the true target item from the remaining candidates in the scene. This requires reasoning about \textit{what} the user truly desires, i.e., the underlying needs and preferences in their expressed utterances.

To address the above two challenges, we introduce \textbf{Si}tuated \textbf{P}ref\textbf{e}rence \textbf{R}easoning (\textbf{\model}), a novel framework that accordingly integrates two key mechanisms. 
First, we present \textbf{scene transition estimation}, which focuses on joint modeling of transition decision and target scene prediction. By leveraging multimodal large language models (MLLMs)~\citep{DBLP:journals/corr/abs-2409-12191,liu2024llavanext} to represent both visual scenes and conversation histories, this mechanism dynamically estimates whether the current scene aligns with user needs, allowing the system to predict a more suitable scene and guide the user to it in the next turn.
Second, considering that LLMs often struggle to disentangle nuanced preferences from surface-level conversation, we formalize preference discovery as a \textbf{Bayesian inverse inference}~\cite{baker2009action, ullman2009help,DBLP:conf/acl/JinWCXKHU0TS24} problem. 
This approach treats the user's utterance as an observable action generated by a latent goal. 
By leveraging two opposing hypothetical beliefs (like vs. dislike), we quantify the likelihood of each potential item being the ``what'' the user desires. 
This allows the system to move beyond heuristic guesses and perform more rigorous probabilistic reasoning.

Our contributions are summarized as follows: 
\begin{itemize}[leftmargin=*, nolistsep]
    \item We identify the \textbf{unique yet underexplored challenges} in situated conversational recommendation (SCR): reasoning \textit{dynamic} and \textit{implicit} user preferences in grounded, evolving scenes. Bridging this gap is important for delivering contextually appropriate recommendations in real-world settings.
    \item In light of these challenges, we propose \textbf{\model}, a novel situated preference reasoning framework that integrates scene transition estimation and Bayesian inverse inference. To our knowledge, this work is among the early framework-level attempts to systematically address SCR.
    \item  Our \model achieves notable improvements over the compared baselines, with an average improvement of 10.9\% on SIMMC 2.1~\citep{kottur2023overview} and 10.6\% on SCREEN~\citep{DBLP:conf/mm/LinWLL24}, respectively. Further analyses validate the effectiveness of each proposed mechanism, providing valuable insights into the development of practical SCR systems.
\end{itemize}

\section{Related Work}

\paragraph{Conversational Recommender Systems.}
Existing conversational recommender systems~\cite{li2018towards,liu2020towards}(CRSs) seek to improve recommendation quality by focusing on two main aspects: learning effective item representations \citep{DBLP:conf/nips/ZhangYSJC19, zhou2020improving} and understanding user preferences conveyed in dialogues \citep{DBLP:conf/sigir/DengL0DL21, DBLP:conf/aaai/LinWL23}. The former involves learning informative item embeddings that can represent items accurately \citep{DBLP:conf/acl/LuBSMCWH21, zhou2022c2}, while the latter focuses on extracting user preferences from the dialogues to enhance personalization \citep{chen2019towards, DBLP:journals/corr/abs-2110-07477}. However, despite the success of these approaches, they primarily focus on text-based interactions, overlooking the visual information of items. In many real-world scenarios, the visual characteristics of items may significantly influence user preferences \citep{DBLP:conf/acl/LongHYHL023}. Additionally, environmental factors, which influence both the context in which recommendations are made and the user’s interaction with items, can significantly affect the quality of recommendations.

\begin{figure*}[t!]
\centering
\includegraphics[width=\textwidth]{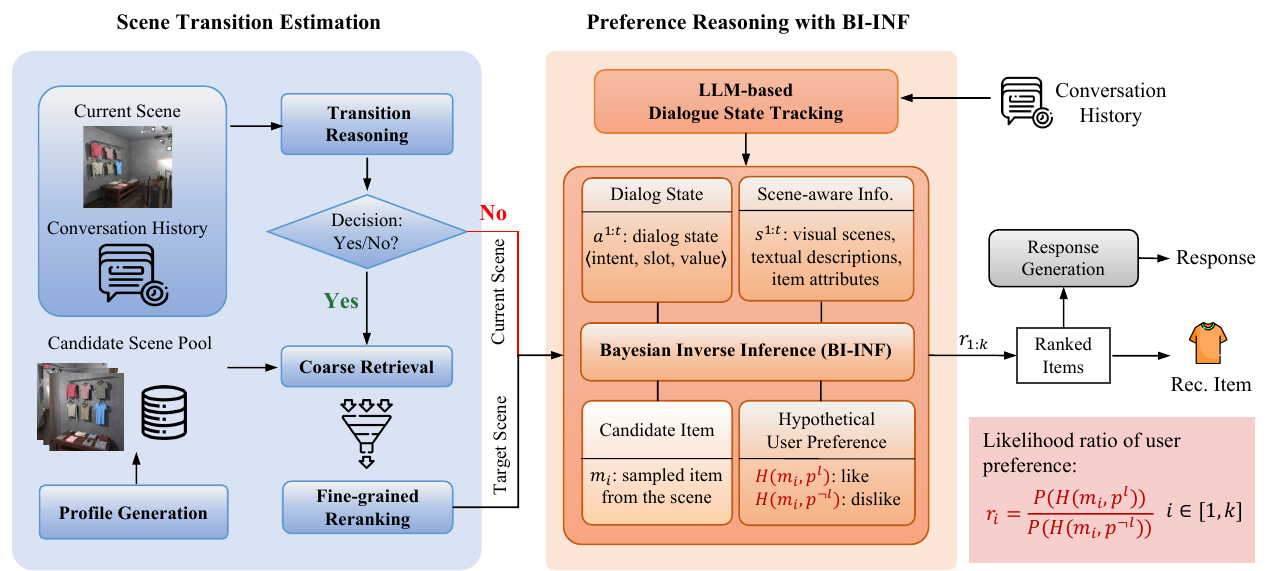}
    \caption{Overview of the Situated Preference Reasoning (\model), which has two critical mechanisms: (a) scene transition estimation (STE) and (b) Bayesian inverse inference (BI-INF).
    }
\label{fig:architecture}
\end{figure*}

\paragraph{Situated Conversational Recommendation.}

In recent years, considerable attention has focused on how user preferences and interests evolve under the influence of situational context~\citep{DBLP:journals/corr/abs-1911-02690}. To advance the development of this emerging field, \citet{DBLP:conf/mm/LinWLL24} pioneered the formalization of situated conversational recommendation (SCR). They leveraged the SIMMC 2.1 dataset~\citep{kottur2023overview} and released the first SCR dataset, SCREEN~\citep{DBLP:conf/mm/LinWLL24}, which significantly contributes to the understanding of dynamic user needs in context-aware conversations. Subsequently, \citet{DBLP:journals/corr/abs-2412-18416} crafted the MUSE dataset by collecting user profiles from real-world scenarios and simulating dialogues using a multi-agent framework powered by MLLMs. These foundational works have significantly contributed to SCR research by providing valuable datasets, enabling more accurate modeling of user behavior and context. Despite these efforts, there is still a lack of comprehensive analysis or a dedicated framework designed to systematically address situated conversational recommendations.
\section{Task Formulation}
Let us consider a shared environment defined by a collection of visual scenes $\{\mathcal{S}_{i}\}_{i=1}^{C}$, where each scene $\mathcal{S}_{i}$ contains a set of candidate items $\{\mathcal{I}_{i,j}\}_{j=1}^{K_i}$, $C$ denotes the total number of scenes in the environment, $K_i$ represents the number of items. These scenes and items are accessible to both users and the recommender assistant (system).
The system engages in multi-turn interactions with a user through natural language conversations, represented as $\{u_t, v_t\}_{t=1}^{T}$, where $u_t$ and $v_t$ denote the user and system utterances at the $t$-th turn, respectively. $T$ denotes the total number of turns.

At each turn $t$, the system operates with (\romannumeral1) a situational context $\mathcal{C}_t$, which includes a specific visual scene $\mathcal{S}_t$ and the corresponding item set $\mathcal{I}_t$ in the scene, and (\romannumeral2) a conversation history $\mathcal{H}_{t}=\{u_{\leq t},v_{<t}\}$, which comprises all past user utterances and system responses. 
The objective of situated conversational recommendation is to generate a contextually appropriate response $v_t$ that adapts to the user's evolving interests. 
This process entails determining the appropriate visual scene (\textit {where}) to ground the conversation and, when suitable, recommending a subset of items (\textit{what}) from the target scene that best satisfy user preferences.
\section{Method}
\label{sec:method}

In this section, we present \textbf{Si}tuated \textbf{P}ref\textbf{e}rence \textbf{R}easoning (\textbf{\model}), a novel framework that comprises two critical mechanisms: scene transition estimation (see \S\ref{sec:scene_transition}) and Bayesian inverse inference (see \S\ref{sec:preference_reasoning}). 
We introduce the end tasks of recommendation and response generation in \S\ref{sec:response_generation}. Figure~\ref{fig:architecture} shows the overview of \model.

\subsection{Scene Transition Estimation}
\label{sec:scene_transition}

To ensure precise and natural scene transitions, we propose a reasoning-driven generative-retrieval framework for Scene Transition Estimation (STE). 
Instead of directly matching the dialogue history to latent scene representations, our framework first externalizes the desired environment as a semantic profile, which provides an explicit anchor for the user's implicit and evolving intent. This generated profile is only an intermediate query rather than the final transition result. We then perform a coarse-to-fine retrieval step to identify the optimal scene from a large-scale candidate pool.

\paragraph{Scene-Profile Representation.} 
Direct reasoning on raw visual scenes is computationally prohibitive and prone to noise. We therefore convert each scene $\mathcal{S}_i$ in the candidate pool into a textual situated profile $d_{\mathcal{S}_i}$, using a pre-trained MLLM. 
Each profile encapsulates the spatial relations and a structured catalog of items with their visual attributes. 
This profile is leveraged to transform target scene prediction into a semantic matching problem.

\paragraph{Profile Generation of Target Scene.}
To maintain proactive yet coherent dialogue, the system must envision a target scene that satisfies the user's implicit intent.
Given the conversation history $\mathcal{H}_t$ and the current scene profile $d_{\mathcal{S}_t}$, we prompt the MLLM $\mathcal{F}$ to perform a joint inference of \textit{transition decision} and \textit{target generation}.
This process aims to determine the necessity of a transition by balancing alignment with user needs against coherence with the current scene.
Specifically, the model is instructed to first generate a decision token $y_{dec} \in \{\text{\texttt{Yes}}, \text{\texttt{No}}\}$, followed by the profile content $d_{\tilde{\mathcal{S}}}$ of the expected target scene:
\begin{equation}
    y_{dec}, d_{\tilde{\mathcal{S}}} = \mathcal{F}(\mathcal{H}_{t}, d_{\mathcal{S}_t}).
\end{equation}
We quantify the likelihood of the transition via the normalized probability of the decision token:
\begin{equation}
    s_{trans} = \frac{\exp(\mathbf{z}_{\text{\texttt{Yes}}})}{\exp(\mathbf{z}_{\text{\texttt{Yes}}}) + \exp(\mathbf{z}_{\text{\texttt{No}}})},
    \label{eq:transition_score}
\end{equation}
where $\mathbf{z}$ denotes the output logits corresponding to the generated tokens. When the transition decision is affirmative (i.e., $y_{dec} = \text{\texttt{Yes}}$), this generated profile $d_{\tilde{\mathcal{S}}}$ serves as a semantic anchor to predict the scene to transit in the retrieval stage. Otherwise, the system retains the current scene $\mathcal{S}_t$ as the grounded context for the downstream stage.

\paragraph{Coarse-to-Fine Transition Reasoning.} 
Since exhaustive semantic reasoning over all candidates is intractable, we employ a coarse-to-fine strategy for transition estimation.
In the \textit{coarse retrieval} stage, we aim to narrow down the scope of the candidate scenes. 
We encode all candidate profiles $\{d_{\mathcal{S}_{i}}\}$ and the reasoned profile $d_{\tilde{\mathcal{S}}}$ into a shared embedding space with an encoder $\phi(\cdot)$. 
The similarity score for each candidate scene is given by:
\begin{equation}
\text{Score}(\mathcal{S}_i) = \frac{\phi(d_{\tilde{\mathcal{S}}}) \cdot \phi(d_{\mathcal{S}_i})}{||\phi(d_{\tilde{\mathcal{S}}})|| \cdot||\phi(d_{\mathcal{S}_{i}})||}.
\end{equation}
We retain the top-$N$ candidates to form a reduced subset $\mathcal{S}_{\text{top}}$ of the scenes.
Then, we take a \textit{fine-grained reranking} for determining the target scene. 
To achieve precise estimation, we train a reranker $f_{\theta}$ (parameterized by an LLM) to evaluate the alignment between $d_{\tilde{\mathcal{S}}}$ and each $d_{\mathcal{S}_{j}}$, where $\mathcal{S}_j \in \mathcal{S}_{\text{top}}$. 
To optimize the reranker, we minimize the following negative log-likelihood during training:
\begin{equation}
\mathcal{L}_{r} = -\log \frac{\exp(f_{\theta}(d_{\tilde{\mathcal{S}}}, d_{\mathcal{S}^{*}}))}{\sum_{\mathcal{S}_{j} \in \mathcal{S}_{\text{top}} \cup \{\mathcal{S}^*\}} \exp(f_{\theta}(d_{\tilde{\mathcal{S}}}, d_{\mathcal{S}_j}))},
\end{equation}
where $d_{\mathcal{S}^{*}}$ denotes the profile of the ground-truth scene. 
This trained reranker ensures that the estimated scene satisfies user needs while maintaining a smooth transition. Consequently, even if the generated profile contains imperfect or partially hallucinated attributes, the final transition decision remains grounded in real candidate scenes rather than unconstrained free-form generation.

\subsection{Bayesian Inverse Inference}
\label{sec:preference_reasoning}
Once the appropriate scene for the next turn is identified, we aim to reason the user's underlying preferences about potential items within that scene. 
Since LLMs often struggle to disentangle nuanced preferences from surface-level conversation, we formalize preference reasoning as a Bayesian inverse inference (BI-INF)~\cite{baker2009action, ullman2009help,DBLP:conf/acl/JinWCXKHU0TS24} problem. 
Our approach consists of the following three stages.

\paragraph{Dialogue State Tracking.}
Dialogue state tracking (DST) aims to estimate the dialogue state at each conversational turn, where the state is typically represented as a set of structured tuples related to system actions or user intents. 
Here, we refer to dialogue states as user intents, such as requesting product information or comparing different items (see Figure \ref{fig:predefined_intent_slot} in the Appendix). 
Specifically, we directly instruct a powerful LLM to extract symbolic dialogue states from dialogue histories in the form of  $\langle$\texttt{intent, slot, value}$\rangle$ tuples. 
To assess reliability, we randomly sampled LLM-extracted states from each downstream dataset and manually verified their correctness against the predefined schema. This approach achieves a high accuracy of 98.8\%, validating the quality of the extracted states.

\paragraph{User Preference Modeling.}
Drawing inspiration from the Bayesian Inverse Planning (BIP) framework used in computational cognitive science~\citep{baker2009action, ullman2009help} and recent multimodal Theory of Mind research~\citep{DBLP:conf/acl/JinWCXKHU0TS24, DBLP:journals/corr/abs-2408-12574}, we approach preference reasoning by \textit{reversing} the user's decision-making process. 
Instead of modeling the system's policy, we formulate the user as a rational agent interacting with the environment.
This process is formalized as a Partially Observable Markov Decision Process (POMDP) defined by the tuple $\langle \mathcal{S}, \mathcal{M}, \mathcal{A}, \mathcal{T}, \pi \rangle$.
Here, $s_t \in \mathcal{S}$ represents the situational context (scene). 
$m_i \in \mathcal{M}$ denotes the user's latent goal (i.e., the target item they desire), and $p_t$ represents their evolving mental state (e.g., beliefs or specific preferences about item attributes).
Crucially, we view the user's utterance as an action $a_t \in \mathcal{A}$ (represented as the dialogue state) taken to achieve their goal.
The user generates these dialogue actions according to a latent policy $\pi(a_t | m_i, p_t, s_t)$, which reflects the likelihood of the user expressing specific intents given their underlying goal $m_i$ and the current context.

Based on this forward generative model, we can infer the user's latent goal $m_i$ by observing their dialogue actions $a_{\leq t}$. 
We represent the posterior probability of the user desiring item $m_i$ as follows:
\begin{align}
    \mathbb{P}(m_i, p_t| & a_{\leq t}, s_{\leq t}) \propto \prod_{\tau=1}^{t} \pi(a_{\tau}|m_i, p_{\tau}) \notag \\ &  \cdot \mathbb{P}(p_\tau|p_{\tau-1},s_\tau) \mathbb{P}(p_0)\mathbb{P}(m_i),
\label{eq:tom_prob}
\end{align}
where $\mathbb{P}(m_i)$ is the prior over items. The term $\mathbb{P}(p_\tau|p_{\tau-1},s_\tau)$ models the dynamics of user preference states. In practice, rather than maintaining an explicit state vector, we approximate this belief update by conditioning the model on the history of dialogue states $a_{<\tau}$ and the situational context~\citep{hausknecht2015deep, rabinowitz2018machine}. The term $\pi(a_{\tau}|m_i, p_{\tau})$ serves as the core \textit{user likelihood} function: it quantifies how likely the user is to produce the dialogue state $a_{\tau}$ if their true goal were item $m_i$.

\paragraph{Inverse Inference through Hypotheses.}
Directly calculating the user policy $\pi(\cdot)$ in Eq. (\ref{eq:tom_prob}) is intractable due to the vast space of natural language. 
To address this, we follow~\citet{DBLP:conf/acl/JinWCXKHU0TS24} to amortize the policy utilizing a fine-tuned MLLM.
This approach leverages the model's world knowledge to simulate the user's behavior.
To infer the user's attitude toward a candidate item $m_i$, we compare two competing hypotheses: 
(\romannumeral1) $\mathcal{H}({m_i,p_t^{l}})$, denoting the user \textit{likes} (or accepts) item $m_i$; and 
(\romannumeral2) $\mathcal{H}({m_i,p_t^{\neg l}})$, denoting the user \textit{dislikes} (or rejects) item $m_i$. 
We compute the likelihood ratio of these hypotheses as:
\begin{align}
    \frac{\mathbb{P}(m_i, p_t^{l})}{\mathbb{P}(m_i, p_t^{\neg l})}&\approx
    \frac{\pi(a_t|m_i,p_t^l)\cdot \mathbb{P}(p_t^l|{p}_{t-1}^{l},s_t)}{\pi(a_t|m_i,p_t^{{\neg}l})\cdot \mathbb{P}(p_t^{{\neg}l}|{p}_{t-1}^{{\neg}l},s_t)}\notag\\
    &\cdot\frac{\prod^{t-1}_{\tau=1}\pi(a_\tau|m_i,\hat{p}_{\tau}^{l})}{\prod^{t-1}_{\tau=1}\pi(a_\tau|m_i,\hat{p}_{\tau}^{{\neg}l})},
\label{eq:likelihood_ratio}
\end{align}
where $\hat{p}_{\tau}$ denotes the estimated belief state derived from the history up to turn $\tau$, the policy $\pi(a_t|m_i, p)$ is approximated by the MLLM's generation probability. 
Specifically, we feed the MLLM with the situational context, the target item $m_i$, and a hypothesis prompt (e.g., ``The user wants this item''). The MLLM then computes the probability of generating the observed dialogue state $a_t$ (e.g., ``Any other options?'').
A higher likelihood under the ``like'' hypothesis compared to the ``dislike'' hypothesis indicates that the observed utterance is more consistent with the user desiring that specific item.
The input-output format during fine-tuning is shown in Figure~\ref{fig:MLLM_Finetune_Template} in the Appendix.

During inference, we calculate the preference ratio $r_i$ for each candidate item $m_i$ in the scene. This is given by: 
\begin{equation}
    r_i=\frac{\mathbb{P}(\mathcal{H}({m_i,p_t^{l}}))}{\mathbb{P}(\mathcal{H}({m_i,p_t^{\neg l}}))}.
\end{equation}
Items with higher ratios are deemed as the user's probable targets and are passed to the system.

\subsection{Recommendation \& Response Generation}
\label{sec:response_generation}
\begin{table*}[t!]
\centering
\resizebox{1\textwidth}{!}{
\begin{tabular}{cl ccccc ccccc}
\toprule
\multirow{2}{*}{\textbf{Type}}  & \multirow{2}{*}{\textbf{Model}} & \multicolumn{5}{c}{\textbf{SIMMC 2.1}} & \multicolumn{5}{c}{\textbf{SCREEN}} \\
\cmidrule(lr){3-7} \cmidrule(lr){8-12}   
&  & \textbf{R@1} & \textbf{R@3} & \textbf{R@5} & \textbf{MRR@3} & \textbf{MRR@5} & \textbf{R@1} & \textbf{R@3} & \textbf{R@5} & \textbf{MRR@3} & \textbf{MRR@5}\\
\midrule
\multirow{4}{*}{CoT} &  LLaVA-NeXT \citep{liu2024llavanext}& 13.01 & 13.92 & 14.12 & 13.45 & 13.52 & 15.42 & 16.85 & 18.21 & 15.68 & 15.98\\
& Qwen2.5-VL \citep{DBLP:journals/corr/abs-2409-12191}& 16.72 & 18.35 & 18.61 & 17.65 & 17.92 & 21.05 & 23.68 & 24.12 & 23.01 & 23.42\\
& GPT-4o \citep{openai2024gpt4o}& 28.12 & 45.42 & 53.18 & 36.21 & 38.05 & 33.45 & 49.32 & 58.15 & 42.21 & 44.58\\
\midrule
\multirow{4}{*}{ICL} & LLaVA-NeXT \citep{liu2024llavanext} & 14.36 & 15.26 & 15.48 & 14.76 & 15.10 & 16.71 & 18.22 & 18.80 & 16.67 & 17.20\\
& Qwen2.5-VL \citep{DBLP:journals/corr/abs-2409-12191}& 17.12 & 19.66 & 20.02 & 19.14 & 19.45 & 21.18 & 23.24 & 23.96 & 22.95 & 23.52 \\
& GPT-4o \citep{openai2024gpt4o} & 29.15 & 47.94 & 55.45 & 38.45 & 39.95 & 35.06 & 49.94 & 60.16 & 44.58 & 45.96\\
\midrule
\multirow{8}{*}{Training} & ALBEF \citep{DBLP:conf/nips/LiSGJXH21}& 6.06 & 7.45 & 8.19 & 7.28 & 7.45 & 8.51 & 9.98 & 12.64 
 & 10.75 & 12.03\\
& LLaVA-NeXT \citep{liu2024llavanext} & 23.67 & 26.84 & 30.18 & 24.89 & 27.77 & 24.63 & 28.49 & 30.46 & 27.98 & 29.11\\
& Qwen2.5-VL \citep{DBLP:journals/corr/abs-2409-12191} & 29.47 & 31.69 & 37.16 & 29.20 & 30.42 & 32.06 & 35.02 & 37.26 & 34.01 & 35.32\\
& ReGeS \citep{yang2025reges} & 27.68 & 45.45 & 54.12 & 35.49 & 37.46 & 31.42 & 49.85 & 59.24 & 39.88 & 41.75 \\
\cmidrule{2-12}
&  \textbf{\model (Ours)} & \bfseries{38.75} & \bfseries{54.09} & \bfseries{58.61} & \bfseries{45.80} & \bfseries{46.83} & \bfseries{39.41} & \bfseries{54.95} & \bfseries{63.80} & \bfseries{50.36} & \bfseries{51.95} \\
& ~~ w/o STE & 33.69 & 47.85 & 52.32 & 40.29 & 41.66 & 30.26 & 43.88 & 51.16 & 40.71 & 42.54\\
& ~~ w/o BI-INF & 31.88 & 44.26 & 47.51 & 38.55 & 39.13 & 33.96 & 48.49 & 51.96 & 46.24 & 47.92 \\
\bottomrule
\end{tabular}}
\caption{Performance of different methods on preference reasoning (recommendations). All results are presented as percentages (\%). The best results per metric are highlighted in bold ($t$-test with $p$-value $<$ 0.05).}
\label{table:recommendation_result}
\vspace{-8pt}
\end{table*}
After ranking all in-scene items based on their inferred preference likelihood ratio, we select the top-$k$ candidates for recommendation.
Following recent advances in generative recommendation~\citep{DBLP:conf/recsys/NieZYDZCZCLCXH24, DBLP:conf/ecir/HouZLLXMZ24}, we employ MLLMs to produce natural, context-aware system responses directly. 
To this end, we concatenate the task-specific instruction, the metadata of the top-$k$ candidate items, the dialogue history, and the description of the target visual scene together as a prompt and feed it into an MLLM to generate the next-turn response. 
By considering both situational and conversational contexts, this approach effectively enhances the relevance of the system's recommendations that satisfy user preferences.

\section{Experiments}

\subsection{Experimental Setup}
\paragraph{Datasets.}
We evaluate our method using two publicly available SCR datasets: \textbf{SIMMC 2.1}~\citep{kottur2023overview} and \textbf{SCREEN}~\citep{DBLP:conf/mm/LinWLL24}. 
The SIMMC 2.1 dataset provides a multimodal, task-oriented dialogue corpus that captures interactions between customers and sales assistants within an immersive 3D virtual shopping environment. The SCREEN dataset comprises over 20,000 synthetic dialogues focused on situated conversational recommendations. 
Appendix~\ref{appendix:datasets} provides dataset statistics and further preprocessing details. In particular, our evaluation split is balanced to include 50\% transition-required dialogues, and over 90\% of SCREEN dialogues require implicit preference refinement beyond the initial user request.

\paragraph{Baseline Methods.}
Since the task of situated conversational recommendation remains underexplored, selecting suitable baseline methods for fair comparison is challenging. To this end, we evaluate representative models across three distinct learning paradigms: 
1) \textbf{Chain-of-Thought (CoT)}: We utilize strong MLLMs in a zero-shot manner, instructing them to reason step-by-step about the visual scene and user intent. This includes the proprietary GPT-4o~\citep{openai2024gpt4o}, as well as open-source LLaVA-NeXT~\citep{liu2024llavanext} and Qwen2.5-VL~\citep{DBLP:journals/corr/abs-2409-12191}. 
2) \textbf{In-Context Learning (ICL)}: To mitigate zero-shot limitations, we enhance these backbones by prepending retrieved, semantically similar dialogue-recommendation demonstrations to the input context. 
3) \textbf{Training-based Methods}: This category comprises fully supervised models, including ALBEF~\citep{DBLP:conf/nips/LiSGJXH21}, a representative small-scale multimodal model, and ReGeS~\citep{yang2025reges}, a specialized text-based generative recommender. 
Unless otherwise noted, all vision-language baselines are provided with the raw scene image, the dialogue history, and the textual item metadata for the current scene. For the text-only ReGeS baseline, we replace raw images with structured scene profiles so that it receives the same environment information in text form.
Regarding optimization, ALBEF undergoes full-parameter fine-tuning, whereas ReGeS and the large-scale MLLM baselines utilize Low-Rank Adaptation (LoRA)~\citep{DBLP:conf/iclr/HuSWALWWC22,DBLP:conf/nips/DettmersPHZ23} for efficient adaptation. Detailed configurations and implementations for all baseline methods are provided in Appendix~\ref{appendix:model_details}.

\paragraph{Implementation Details.}
We adopt Qwen2.5-VL-7B-Instruct as the core MLLM for the \model framework. GPT-4o is used only in an offline preprocessing stage for scene captioning and profile generation, and is not queried during online turn-by-turn inference.
In the STE module, we employ Qwen3-Embedding-4B as the dense encoder $\phi(\cdot)$ for coarse retrieval and Qwen3-Reranker-4B as the backbone for the fine-grained reranker $f_{\theta}$.
In the BI-INF module, we amortize the Bayesian policy $\pi(\cdot)$ by fine-tuning the Qwen2.5-VL backbone to predict the structured state $a_t$ via cross-entropy loss. At inference, the policy probability is computed from the output logits of the observed structured state, rather than by autoregressively generating a full response for every candidate item.
We optimize the model using AdamW~\citep{DBLP:conf/iclr/LoshchilovH19} and employ nucleus sampling~\citep{holtzman2020curious} for response generation.
Detailed hyperparameters for model architecture, training, and generation are listed in Appendix~\ref{appendix:implementation_details}. All prompting templates used are provided in Appendix~\ref{appendix:prompt}. 

\paragraph{Efficiency Considerations.}
Our framework is designed to keep the online deployment cost manageable. First, the only proprietary component, GPT-4o, is used once offline for scene-profile construction and is not involved in turn-by-turn inference. Second, BI-INF does not autoregressively generate a complete response for every candidate item; instead, it scores the already observed dialogue state directly from model logits and invokes response generation only after candidate ranking. The detailed latency breakdown, scene-density scaling analysis, and the remaining discussions are reported in Appendix~\ref{appendix:additional_analyses}.
Empirically, \model requires a similar time cost compared with the strongly trained Qwen2.5-VL baseline, while improving R@1 from 29.47 to 38.75; its latency also scales roughly linearly from $\sim$0.8s to $\sim$2.9s as the number of in-scene items increases (Tables~\ref{tab:efficiency_breakdown} and~\ref{tab:latency_bucket}).

\begin{table*}[t!]
\centering
 \resizebox{1\textwidth}{!}{
 \begin{tabular}{cl ccc ccc}
 \toprule
\multirow{2}{*}{\textbf{Type}}  & \multirow{2}{*}{\textbf{Model}} & \multicolumn{3}{c}{\textbf{SIMMC 2.1}} & \multicolumn{3}{c}{\textbf{SCREEN}} \\
 \cmidrule(lr){3-5} \cmidrule(lr){6-8}  
 & & \textbf{BLEU-1/2} & \textbf{ROUGE-1/L} & \textbf{GPT-Score} & \textbf{BLEU-1/2} & \textbf{ROUGE-1/L} & \textbf{GPT-Score} \\
\midrule
\multirow{4}{*}{CoT} & LLaVA-NeXT \citep{liu2024llavanext} & 18.42 / 8.51 & 15.62 / 12.58 & 5.64 & 28.75 / 19.68 & 23.85 / 18.24 & 6.12 \\
& Qwen2.5-VL \citep{DBLP:journals/corr/abs-2409-12191} & 19.85 / 10.64 & 15.88 / 12.72 & 5.85 & 34.22 / 22.85 & 25.56 / 19.12 & 6.25 \\
& GPT-4o \citep{openai2024gpt4o} & 30.24 / 14.52 & 20.85 / 18.62 & 7.24 & 40.52 / 28.34 & 40.72 / 36.21 & 7.85 \\
 \midrule
 \multirow{4}{*}{ICL} & LLaVA-NeXT \citep{liu2024llavanext} & 21.89 / 10.15 & 18.67 / 15.55 & 5.92 & 33.26 / 22.90 & 30.59 / 23.21 & 6.42 \\
 & Qwen2.5-VL \citep{DBLP:journals/corr/abs-2409-12191} & 22.28 / 11.30 & 19.30 / 16.86 & 6.15 & 37.75 / 26.34 & 31.58 / 24.92 & 6.58 \\
 & GPT-4o \citep{openai2024gpt4o} & 27.70 / 13.92 & 21.69 / 20.52 & 7.56 & 42.39 / 31.56 & 42.04 / 36.11 & 8.12 \\
 \midrule
 \multirow{7}{*}{Training} &  ALBEF \citep{DBLP:conf/nips/LiSGJXH21} & 21.65 / 10.18 & 17.02 / 15.41 & 6.75 & 34.29 / 24.12 & 26.21 / 20.23 & 6.92 \\
 & LLaVA-NeXT \citep{liu2024llavanext} & 27.13 / 18.88 & 22.89 / 20.25 & 7.82 & 43.67 / 29.92 & 38.29 / 33.29 & 8.24 \\
 & Qwen2.5-VL \citep{DBLP:journals/corr/abs-2409-12191}& 29.77 / 19.31 & 24.87 / 21.91 & 8.05 & 45.34 / 33.90 & 40.77 / 35.91 & 8.52 \\
 & ReGeS \citep{yang2025reges} & 23.64 / 19.78 & 22.61 / 19.61 & 7.52 & 39.18 / 31.52 & 37.45 / 33.12 & 8.08 \\
 \cmidrule{2-8}
 & \textbf{\model (Ours)} & \bfseries{33.77 / 21.67} & \bfseries{32.61 / 25.52} & \bfseries{8.92} & \bfseries{49.50 / 36.44} & \bfseries{45.48 / 38.50} & \bfseries{9.35}\\
 & ~~ w/o STE & 30.28 / 19.45 & 28.21 / 22.99 & 8.45 & 46.22 / 34.45 & 41.08 / 36.21 & 8.78\\
 & ~~ w/o BI-INF & 31.32 / 19.88 & 29.31 / 23.14 & 8.62  & 47.42 / 35.21 & 41.88 / 37.29 & 8.95 \\
 \bottomrule
 \end{tabular}}
\caption{Performance of different methods on response generation. The best results per metric are highlighted in bold ($t$-test with $p$-value $<$ 0.05).
}
\label{table:generation_result}
\vspace{-8pt}
\end{table*}

\paragraph{Evaluation Metrics.}
We evaluate the performance of SCR models from two aspects: preference reasoning accuracy and response generation quality.
For preference reasoning, we adopt standard metrics for recommendation evaluation: Recall@$k$ (\textbf{R@$k$}, where $k$ = 1, 3, 5) and Mean Reciprocal Rank@$k$ (\textbf{MRR@$k$}, where $k$ = 3, 5). These metrics assess the model’s ability to rank the ground-truth items among the top-$k$ candidates.
For response generation, we conduct both automatic and human evaluations. The automatic evaluation relies on \textbf{BLEU}-1,2 \citep{papineni2002bleu} and \textbf{ROUGE}-1,L \citep{lin2004rouge}, which measure the lexical overlap between generated and reference responses.
To assess semantic coherence and relevance, we additionally employ GPT-4o as a judge to automatically score the generated responses on a scale of 1 $\sim$ 10 (\textbf{GPT-Score}), following established protocols \citep{wang-etal-2024-instruct}. The specific prompting template for GPT-Score is provided in Appendix~\ref{appendix:gpt_score}. 
Details on human evaluation are provided in \S\ref{sec:human_eval}.

\subsection{Main Results}

\paragraph{Can our method achieve effective preference reasoning?}

Table~\ref{table:recommendation_result} details the performance of preference reasoning.
\model achieves the strongest overall recommendation performance among the compared baselines, outperforming proprietary models such as GPT-4o and specialized training-based methods.
Notably, regarding R@1, \model surpasses the second-best Qwen2.5-VL by a margin of 9.28\% on SIMMC 2.1.
Moreover, our framework outperforms ReGeS, a competitive text-based recommender.
This performance gap validates two findings. First, visual information is indispensable for situated recommendation since text-only models fail to capture visual-dependent preferences. Second, our Bayesian inverse inference mechanism is more effective at uncovering implicit user intents than the standard CoT reasoning used in baseline MLLMs.

\paragraph{How does our preference reasoning affect response generation?}

Accurate preference reasoning is the cornerstone of generating appropriate responses.
As shown in Table~\ref{table:generation_result}, \model achieves the strongest overall response-generation performance among the compared methods.
An important observation is that \model achieves higher GPT-Scores (8.92 vs. 7.56 on SIMMC 2.1) than GPT-4o despite using a smaller 7B backbone.
This suggests that general linguistic fluency alone is insufficient for SCR.
By integrating precise scene estimation and preference inference, our method ensures generated responses are not only natural but also factually aligned with latent user needs.

\subsection{Ablation Study}

\begin{figure}[t!]
    \centering
    \begin{subfigure}[t]{0.49\linewidth}
        \centering
        \includegraphics[width=\linewidth]{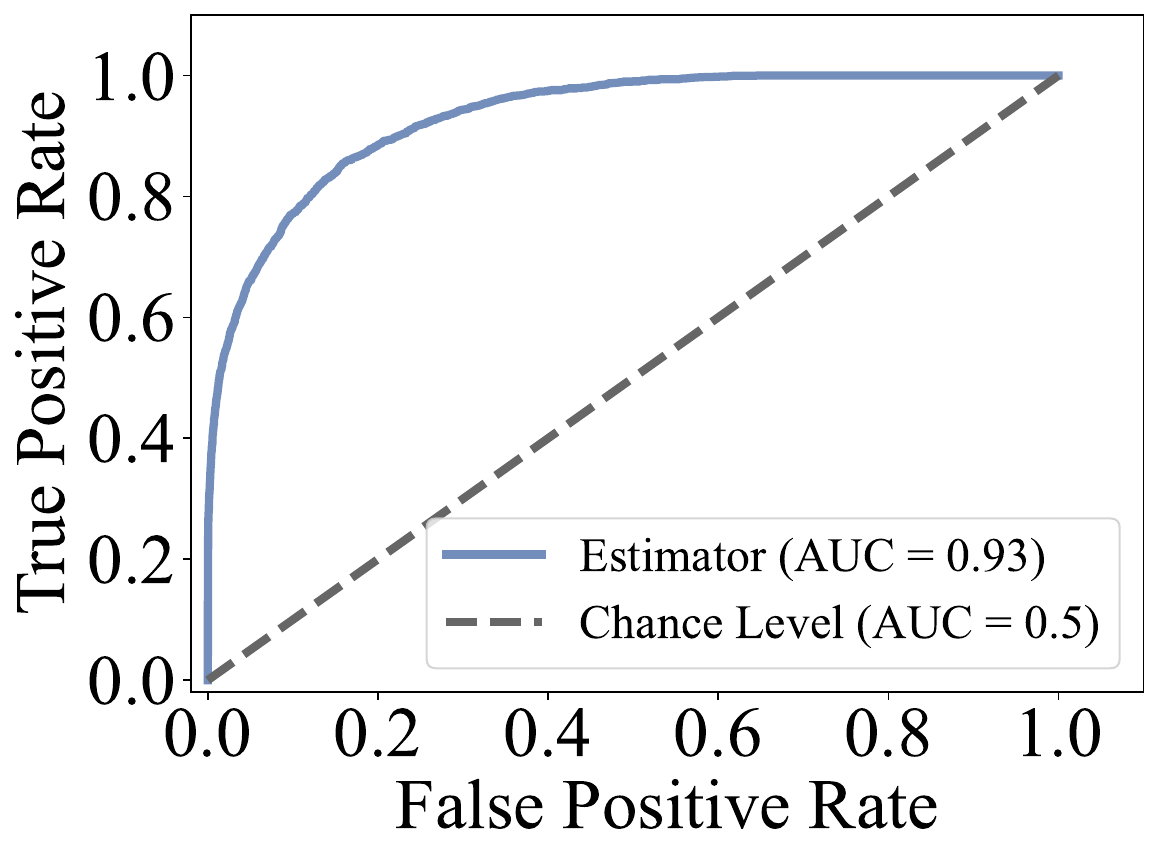}
        \caption{ROC Curve}
        \label{fig:roc_curve}
    \end{subfigure}
    \hfill
    \begin{subfigure}[t]{0.49\linewidth}
        \centering
        \includegraphics[width=\linewidth]{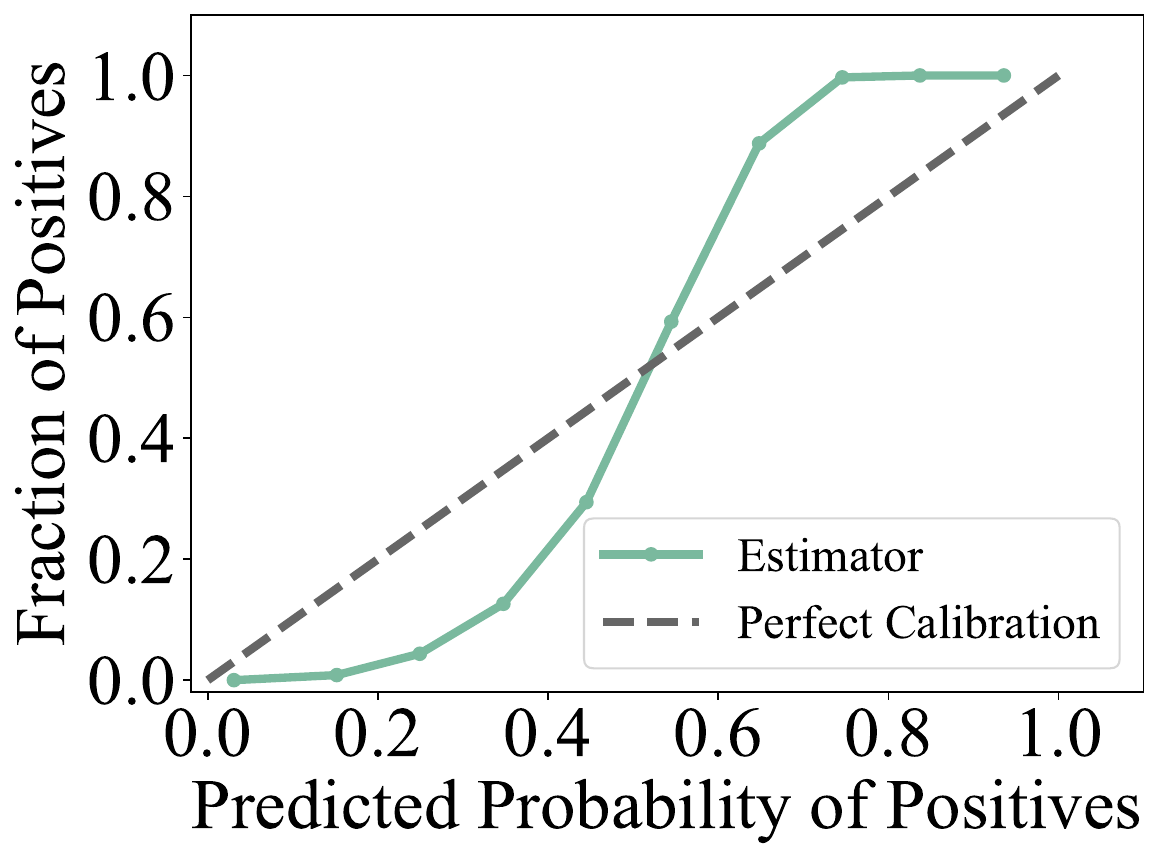}
        \caption{Calibration Curve}
        \label{fig:calibration_curve}
    \end{subfigure}
    \caption{Performance of (a) ROC curve and (b) calibration curve for the scene transition estimation.}
    \vspace{-5pt}
    \label{fig:scene_transition_estimator_analysis}
\end{figure}

\begin{figure}[t!]
    \centering
    \includegraphics[width=1\linewidth]{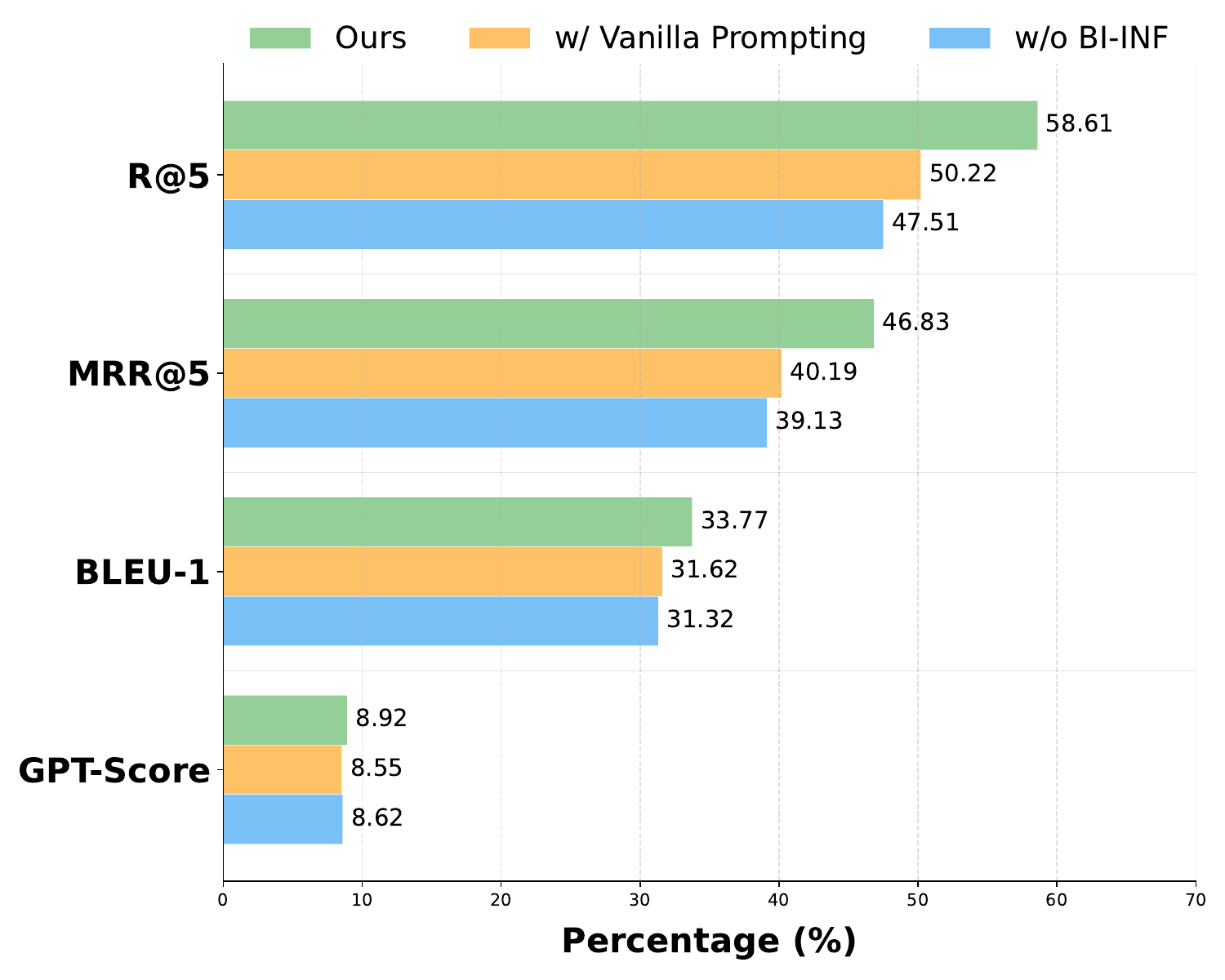}
    \caption{Comparison of different preference inference variants in our \model~on the SIMMC 2.1 dataset.}
    \label{fig:compare_prompt_tom}
    \vspace{-8pt}
\end{figure}

To validate the effectiveness of the proposed mechanisms, we analyze the performance variants by removing one component at a time. The corresponding recommendation and response-generation results are reported in Tables~\ref{table:recommendation_result} and~\ref{table:generation_result}, respectively. We further provide STE-specific analyses in Figure~\ref{fig:scene_transition_estimator_analysis} and Table~\ref{tab:robustness_transitions}, and a BI-INF-specific comparison in Figure~\ref{fig:compare_prompt_tom}.
1) \textbf{without (w/o) STE}: Removing STE leads to the most significant degradation across all metrics. For instance, R@1 drops sharply from 39.41\% to 30.26\% on SCREEN. This decline occurs because, without STE, the system fails to navigate to the correct visual environment. Consequently, the agent remains confined to the irrelevant scene, making it impossible to recommend the correct items or generate responses that align with the user's new intent.  
2) \textbf{without (w/o) BI-INF}: Removing this module results in a considerable performance drop, particularly in ranking metrics like MRR@5. This result highlights that standard MLLM prompting is insufficient for distinguishing the user's true target from visual distractors. BI-INF effectively bridges this gap by rigorously quantifying the likelihood of user utterances, ensuring that the system accurately identifies specific items to recommend.

\begin{table}[t]
\centering
\resizebox{\linewidth}{!}{
\begin{tabular}{lccc ccc}
\toprule
\multirow{2}{*}{\textbf{Method}} & \multicolumn{3}{c}{\textbf{SIMMC 2.1}} & \multicolumn{3}{c}{\textbf{SCREEN}}\\
 \cmidrule(lr){2-4}\cmidrule(lr){5-7}
 & R@1 & MRR@3 & MRR@5 & R@1 & MRR@3 & MRR@5 \\
\midrule
\model (Ours)   & 38.75 & 45.80 & 46.83 & 39.41 & 50.36 & 51.95\\
\midrule
w/ Random      & 34.17 & 40.61 & 42.08 & 31.84 & 41.79 & 43.13 \\
w/ Non-target  & 33.92 & 40.43 & 41.92 & 31.13 & 41.23 & 42.77 \\
w/o STE        & 33.69 & 40.29 & 41.66 & 30.26 & 40.71 & 42.54 \\
\bottomrule
\end{tabular}}
\caption{Comparison of different scene transition estimation variants in \model.}
\label{tab:robustness_transitions}
\vspace{-8pt}
\end{table}

\subsection{Impact of Scene Transition Estimation}
To validate the efficacy of STE, we first evaluate its sub-components. Notably, the generated target profile only serves as an intermediate semantic query; the final transition target is grounded by coarse-to-fine matching over the real candidate pool, which helps buffer occasional profile-generation noise. The transition decision module achieves an AUC of 0.93 with strong calibration (Figure~\ref{fig:scene_transition_estimator_analysis}), and the target scene predictor remains reliable across both datasets in our evaluation. Furthermore, we differentiate the gain of STE from mere architectural complexity by comparing it with randomized (\textit{w/ Random}) and erroneous (\textit{w/ Non-target}) transition strategies. As shown in Table~\ref{tab:robustness_transitions}, both variants suffer significant performance drops (e.g., -4.6\% to -7.6\% in R@1) compared to \model, yet remain superior to the complete removal of STE. This confirms that the performance gains stem specifically from the precise estimation of scene transitions, rather than from merely introducing additional decision steps.

To isolate the contribution of the generative-retrieval design, we additionally include a Qwen3-based retrieval-only transition baseline that directly retrieves target scenes from dialogue history using the identical Qwen3-Embedding-4B retriever as STE, but without explicit target-profile generation. We further compare against a global-access Qwen2.5-VL baseline that receives the top-5 retrieved scene candidates, and provide a conditioned error-propagation analysis together with a dedicated discussion of STE boundary cases in Appendix~\ref{appendix:additional_analyses}. These complementary analyses clarify both the value and the failure modes of the transition module.
Concretely, the retrieval-only variant reaches 35.24 R@1, confirming that direct semantic matching is weaker than our explicit target-profile reasoning, while a global-access Qwen2.5-VL baseline given the top-5 retrieved scenes improves only to 19.58 R@1 and incurs a 138.2\% latency increase (Tables~\ref{tab:retrieval_only_baseline} and~\ref{tab:global_access_baseline}).
Moreover, when conditioning on STE correctness, downstream recommendation quality drops from 40.0 to 29.8 R@1 once the predicted scene is wrong, directly confirming the error-propagation effect from scene grounding to BI-INF (Table~\ref{tab:error_propagation_conditioned}).

\subsection{Impact of Bayesian Inverse Inference}
We further isolate the contribution of the BI-INF module by comparing it against two variants: (a) \textit{w/o BI-INF}, and (b) \textit{w/ Vanilla Prompting}, which replaces the Bayesian framework with direct MLLM instructions. As illustrated in Figure~\ref{fig:compare_prompt_tom}, \model consistently outperforms the vanilla prompting baseline across metrics (e.g., significant gains in R@5 and MRR@5). This shows that rigorous probabilistic reasoning offers a more robust mechanism for disentangling user intent than heuristic single-pass generation, while the gap between vanilla prompting and w/o BI-INF validates that explicit preference modeling is essential for SCR.

\subsection{Human Evaluation and Case Study}
\label{sec:human_eval}

\begin{figure}[t!]
    \centering
    \includegraphics[width=1\linewidth]{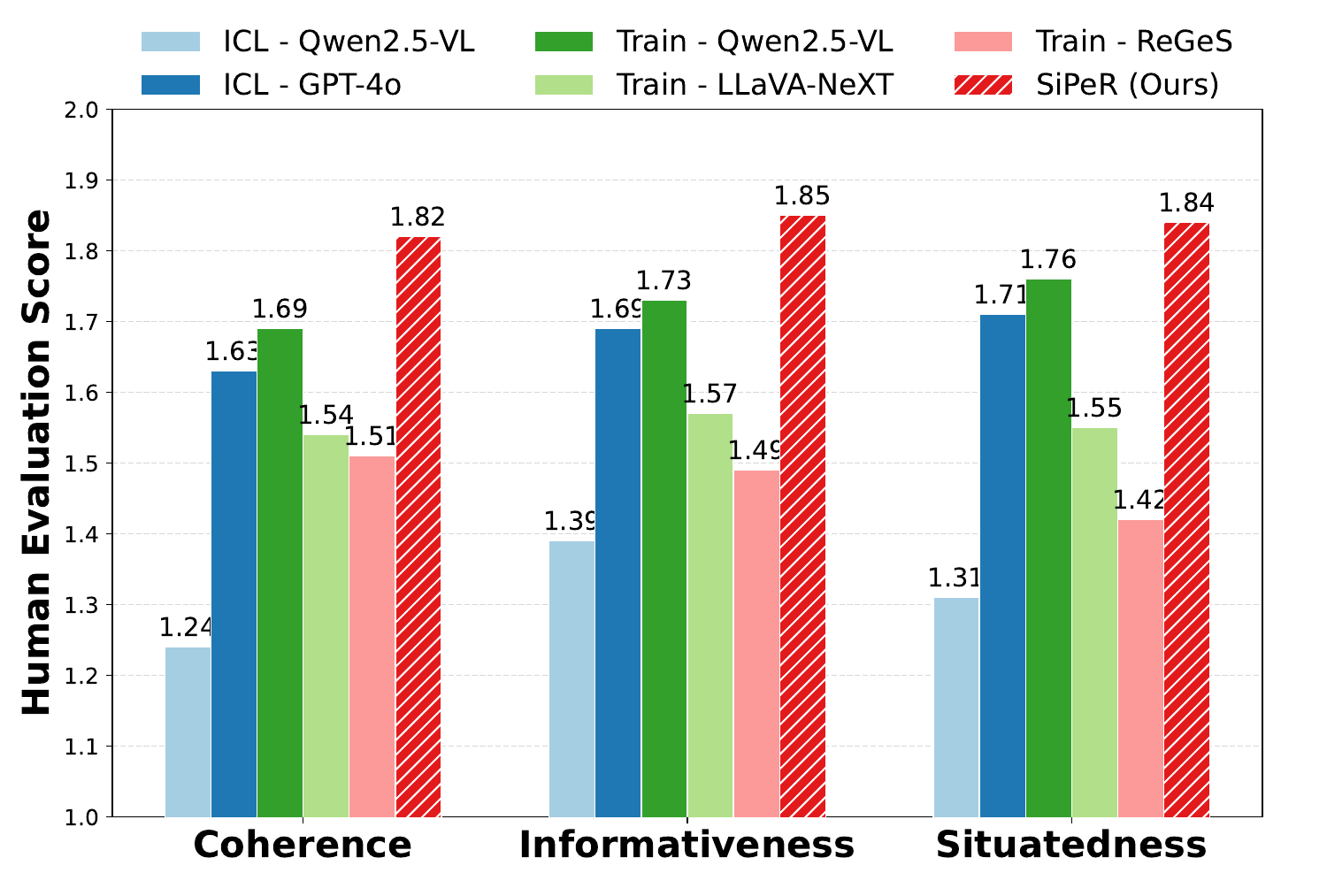} 
    \caption{
        Human evaluation results comparing \model with baselines across Coherence, Informativeness, and Situatedness. The inter-annotator agreement was moderate across all metrics (Fleiss' $\kappa \in [0.47, 0.53]$).
    }
    \label{fig:human_eval_radar}
    \vspace{-8pt}
\end{figure}

We conducted a human evaluation on 30 randomly sampled instances from the SCREEN dataset. Three well-educated annotators independently and blindly rated responses on \textit{Coherence}, \textit{Informativeness}, and \textit{Situatedness} (details in Appendix~\ref{appendix:human_evaluation}).
As shown in Figure~\ref{fig:human_eval_radar}, \model consistently demonstrates comprehensive superiority over all baselines across these dimensions. Notably, it achieves substantial gains in \textit{Situatedness} (1.84) compared to strong baselines such as GPT-4o (1.71) and ReGeS (1.42). We attribute this success to our structured framework: the scene transition estimation ensures the system operates within the correct visual context, while Bayesian inverse inference effectively filters out irrelevant items to target the user's true intent. Consequently, \model generates responses that are not only linguistically fluent but also visually faithful, whereas text-only models like ReGeS struggle significantly without visual cues. The reliability of these results is supported by moderate inter-annotator agreement (Fleiss' $\kappa \in [0.47, 0.53]$)~\citep{fleiss1971measuring}.
To demonstrate the qualitative performance of different models, we selected representative cases from the SIMMC 2.1 test set and presented them in Appendix~\ref{appendix:case_study} for a more detailed examination.
\section{Conclusion}
In this paper, we introduce Situated Preference Reasoning (\model), a novel framework designed to address the challenges of SCR: specifically, reasoning about dynamic scene transitions and implicit user intents. By integrating scene transition estimation and Bayesian inverse inference, \model not only determines \textit{where} to ground the conversation but also infers \textit{what} the user desires through probabilistic modeling. Extensive experiments show that \model achieves superior performance over the compared baselines in both recommendation accuracy and response quality. We hope this work offers a constructive foundation for future research on more proactive and context-aware situated recommendation agents.

\section*{Limitations}
While the proposed \model method demonstrates strong performance, we also identify several limitations. First, as the number of candidate items within a scene increases, the computational cost of preference scoring correspondingly grows, and the retrieval overhead also increases. In practice, lightweight coarse filtering may help reduce this cost before applying the full BI-INF procedure. Second, both STE and BI-INF inherit calibration and hallucination risks from the underlying MLLMs. Although STE grounds the final transition target in a real candidate pool through coarse-to-fine retrieval, noisy target-profile generation can still propagate downstream when an incorrect scene is selected. Possible mitigation strategies include inventory-constrained decoding, self-consistency checks, and stronger uncertainty estimation. We leave these directions to future work.

\section*{Ethics Statement}
The MLLMs utilized in this work include both open-source models and closed-source APIs, and our use strictly follows established academic protocols. In particular, GPT-4o is only used for one-time offline preprocessing, while all online components rely on publicly available open-source models. To mitigate potential biases in the recommendation generation process, we have prioritized the diversity and fairness of the datasets used for both training and evaluation. Furthermore, since the system proactively facilitates scene transitions and provides product recommendations, it is crucial to ensure that it does not manipulate or unduly influence the user's decision-making. 
Finally, while AI assistants (e.g., Cursor and ChatGPT) were partially utilized for coding and linguistic refinement, we affirm that all core content and findings in this paper are the original work of the authors.

\section*{Acknowledgments}
This work was supported by the General Research Fund (GRF) of the Research Grants Council of Hong Kong (PolyU 15207122 and PolyU 15205325), and also in part by the PolyU Postdoc Matching Fund Scheme (4-W40Z). The authors would like to thank the anonymous reviewers for their valuable feedback and constructive suggestions.

\bibliography{custom}

@article{DBLP:journals/corr/abs-1911-02690,
  author       = {Paul A. Crook and
                  Shivani Poddar and
                  Ankita De and
                  Semir Shafi and
                  David Whitney and
                  Alborz Geramifard and
                  Rajen Subba},
  title        = {{SIMMC:} Situated Interactive Multi-Modal Conversational Data Collection
                  And Evaluation Platform},
  journal      = {CoRR},
  volume       = {abs/1911.02690},
  year         = {2019},
  eprinttype    = {arXiv},
  eprint       = {1911.02690},
}

@article{jannach2021survey,
  author    = {Dietmar Jannach and
               Ahtsham Manzoor and
               Wanling Cai and
               Li Chen},
  title     = {A Survey on Conversational Recommender Systems},
  journal   = {{ACM} Comput. Surv.},
  volume    = {54},
  number    = {5},
  pages     = {105:1--105:36},
  year      = {2021},
  doi       = {10.1145/3453154},
}

@inproceedings{li2018towards,
  author    = {Raymond Li and
               Samira Ebrahimi Kahou and
               Hannes Schulz and
               Vincent Michalski and
               Laurent Charlin and
               Chris Pal},
  title     = {Towards Deep Conversational Recommendations},
  booktitle = {Advances in Neural Information Processing Systems},
  pages     = {9748--9758},
  year      = {2018},
}

@inproceedings{liu2020towards,
  author    = {Zeming Liu and
               Haifeng Wang and
               Zheng{-}Yu Niu and
               Hua Wu and
               Wanxiang Che and
               Ting Liu},
  title     = {Towards Conversational Recommendation over Multi-Type Dialogs},
  booktitle = {Proceedings of the 58th Annual Meeting of the Association for Computational Linguistics (ACL)},
  pages     = {1036--1049},
  year      = {2020},
  doi       = {10.18653/v1/2020.acl-main.98},
}

@inproceedings{DBLP:conf/acl/LongHYHL023,
  author       = {Yuxing Long and
                  Binyuan Hui and
                  Caixia Yuan and
                  Fei Huang and
                  Yongbin Li and
                  Xiaojie Wang},
  editor       = {Anna Rogers and
                  Jordan L. Boyd{-}Graber and
                  Naoaki Okazaki},
  title        = {Multimodal Recommendation Dialog with Subjective Preference: {A} New
                  Challenge and Benchmark},
  booktitle    = {Findings of the Association for Computational Linguistics: {ACL} 2023,
                  Toronto, Canada, July 9-14, 2023},
  pages        = {3515--3533},
  publisher    = {Association for Computational Linguistics},
  year         = {2023},
}

@inproceedings{papineni2002bleu,
  author    = {Kishore Papineni and
               Salim Roukos and
               Todd Ward and
               Wei{-}Jing Zhu},
  title     = {Bleu: a Method for Automatic Evaluation of Machine Translation},
  booktitle = {Proceedings of the 40th Annual Meeting of the Association for Computational Linguistics (ACL)},
  pages     = {311--318},
  year      = {2002},
  doi       = {10.3115/1073083.1073135},
}

@inproceedings{chen2019towards,
  author    = {Qibin Chen and
               Junyang Lin and
               Yichang Zhang and
               Ming Ding and
               Yukuo Cen and
               Hongxia Yang and
               Jie Tang},
  title     = {Towards Knowledge-Based Recommender Dialog System},
  booktitle = {Proceedings of the 2019 Conference on Empirical Methods in Natural Language Processing and the 9th International Joint Conference on Natural Language Processing, {EMNLP-IJCNLP}},
  pages     = {1803--1813},
  year      = {2019},
  doi       = {10.18653/v1/D19-1189},
}

@article{gao2021advances,
  author    = {Chongming Gao and
               Wenqiang Lei and
               Xiangnan He and
               Maarten de Rijke and
               Tat{-}Seng Chua},
  title     = {Advances and challenges in conversational recommender systems: {A} survey},
  journal   = {{AI} Open},
  volume    = {2},
  pages     = {100--126},
  year      = {2021},
  doi       = {10.1016/j.aiopen.2021.06.002},
}

@inproceedings{zhou2022c2,
  author    = {Yuanhang Zhou and
               Kun Zhou and
               Wayne Xin Zhao and
               Cheng Wang and
               Peng Jiang and
               He Hu},
  title     = {C{\({^2}\)}-CRS: Coarse-to-Fine Contrastive Learning for Conversational Recommender System},
  booktitle = {{WSDM} '22: The Fifteenth {ACM} International Conference on Web Search and Data Mining},
  pages     = {1488--1496},
  year      = {2022},
  doi       = {10.1145/3488560.3498514},
}

@inproceedings{DBLP:conf/aaai/LinWL23,
  author       = {Dongding Lin and
                  Jian Wang and
                  Wenjie Li},
  editor       = {Brian Williams and
                  Yiling Chen and
                  Jennifer Neville},
  title        = {{COLA:} Improving Conversational Recommender Systems by Collaborative
                  Augmentation},
  booktitle    = {Thirty-Seventh {AAAI} Conference on Artificial Intelligence, {AAAI}
                  2023, Thirty-Fifth Conference on Innovative Applications of Artificial
                  Intelligence, {IAAI} 2023, Thirteenth Symposium on Educational Advances
                  in Artificial Intelligence, {EAAI} 2023, Washington, DC, USA, February
                  7-14, 2023},
  pages        = {4462--4470},
  publisher    = {{AAAI} Press},
  year         = {2023},
  url          = {https://doi.org/10.1609/aaai.v37i4.25567},
  doi          = {10.1609/AAAI.V37I4.25567},
}

@inproceedings{zhou2020improving,
  author    = {Kun Zhou and
               Wayne Xin Zhao and
               Shuqing Bian and
               Yuanhang Zhou and
               Ji{-}Rong Wen and
               Jingsong Yu},
  title     = {Improving Conversational Recommender Systems via Knowledge Graph based Semantic Fusion},
  booktitle = {{KDD} '20: The 26th {ACM} {SIGKDD} Conference on Knowledge Discovery and Data Mining},
  pages     = {1006--1014},
  year      = {2020},
  doi       = {10.1145/3394486.3403143},
}

@article{fleiss1971measuring,
  title={Measuring nominal scale agreement among many raters.},
  author={Fleiss, Joseph L},
  journal={Psychological bulletin},
  volume={76},
  number={5},
  pages={378},
  year={1971},
  publisher={American Psychological Association}
}

@inproceedings{DBLP:conf/coling/MoonKCDPLWDBCSG20,
  author       = {Seungwhan Moon and
                  Satwik Kottur and
                  Paul A. Crook and
                  Ankita De and
                  Shivani Poddar and
                  Theodore Levin and
                  David Whitney and
                  Daniel Difranco and
                  Ahmad Beirami and
                  Eunjoon Cho and
                  Rajen Subba and
                  Alborz Geramifard},
  editor       = {Donia Scott and
                  N{\'{u}}ria Bel and
                  Chengqing Zong},
  title        = {Situated and Interactive Multimodal Conversations},
  booktitle    = {Proceedings of the 28th International Conference on Computational
                  Linguistics, {COLING} 2020, Barcelona, Spain (Online), December 8-13,
                  2020},
  pages        = {1103--1121},
  publisher    = {International Committee on Computational Linguistics},
  year         = {2020},
  url          = {https://doi.org/10.18653/v1/2020.coling-main.96},
  doi          = {10.18653/V1/2020.COLING-MAIN.96},
}

@inproceedings{kottur2023overview,
  title={Overview of Situated and Interactive Multimodal Conversations (SIMMC) 2.1 Track at DSTC 11},
  author={Kottur, Satwik and Moon, Seungwhan},
  booktitle={Proceedings of The Eleventh Dialog System Technology Challenge},
  pages={235--241},
  year={2023}
}

@misc{openai2024gpt4o,
  title = {Hello {GPT-4o}},
  author = {OpenAI},
  journal = {OpenAI Blog},
  year = {2024},
  month = {May},
  howpublished = {\url{https://openai.com/index/hello-gpt-4o/}}
}

@inproceedings{DBLP:conf/nips/LiSGJXH21,
  author       = {Junnan Li and
                  Ramprasaath R. Selvaraju and
                  Akhilesh Gotmare and
                  Shafiq R. Joty and
                  Caiming Xiong and
                  Steven Chu{-}Hong Hoi},
  title        = {Align before Fuse: Vision and Language Representation Learning with
                  Momentum Distillation},
  booktitle    = {Advances in Neural Information Processing Systems 34: Annual Conference
                  on Neural Information Processing Systems 2021, NeurIPS 2021, December
                  6-14, 2021, virtual},
  pages        = {9694--9705},
  year         = {2021},
  bibsource    = {dblp computer science bibliography, https://dblp.org}
}

@inproceedings{DBLP:conf/mm/LinWLL24,
  author       = {Dongding Lin and
                  Jian Wang and
                  Chak Tou Leong and
                  Wenjie Li},
  title        = {{SCREEN:} {A} Benchmark for Situated Conversational Recommendation},
  booktitle    = {Proceedings of the 32nd {ACM} International Conference on Multimedia,
                  {MM} 2024, Melbourne, VIC, Australia, 28 October 2024 - 1 November
                  2024},
  pages        = {9591--9600},
  publisher    = {{ACM}},
  year         = {2024},
  url          = {https://doi.org/10.1145/3664647.3681651},
  doi          = {10.1145/3664647.3681651},
}

@inproceedings{DBLP:conf/acl/JinWCXKHU0TS24,
  author       = {Chuanyang Jin and
                  Yutong Wu and
                  Jing Cao and
                  Jiannan Xiang and
                  Yen{-}Ling Kuo and
                  Zhiting Hu and
                  Tomer D. Ullman and
                  Antonio Torralba and
                  Joshua B. Tenenbaum and
                  Tianmin Shu},
  title        = {MMToM-QA: Multimodal Theory of Mind Question Answering},
  booktitle    = {Proceedings of the 62nd Annual Meeting of the Association for Computational
                  Linguistics (Volume 1: Long Papers), {ACL} 2024, Bangkok, Thailand,
                  August 11-16, 2024},
  pages        = {16077--16102},
  publisher    = {Association for Computational Linguistics},
  year         = {2024},
  url          = {https://doi.org/10.18653/v1/2024.acl-long.851},
  doi          = {10.18653/V1/2024.ACL-LONG.851},
}

@article{DBLP:journals/corr/abs-2409-12191,
  author       = {Peng Wang and
                  Shuai Bai and
                  Sinan Tan and
                  Shijie Wang and
                  Zhihao Fan and
                  Jinze Bai and
                  Keqin Chen and
                  Xuejing Liu and
                  Jialin Wang and
                  Wenbin Ge and
                  Yang Fan and
                  Kai Dang and
                  Mengfei Du and
                  Xuancheng Ren and
                  Rui Men and
                  Dayiheng Liu and
                  Chang Zhou and
                  Jingren Zhou and
                  Junyang Lin},
  title        = {Qwen2-VL: Enhancing Vision-Language Model's Perception of the
                  World at Any Resolution},
  journal      = {CoRR},
  volume       = {abs/2409.12191},
  year         = {2024},
  url          = {https://doi.org/10.48550/arXiv.2409.12191},
  doi          = {10.48550/ARXIV.2409.12191},
}

@inproceedings{DBLP:conf/iclr/HuSWALWWC22,
  author       = {Edward J. Hu and
                  Yelong Shen and
                  Phillip Wallis and
                  Zeyuan Allen{-}Zhu and
                  Yuanzhi Li and
                  Shean Wang and
                  Lu Wang and
                  Weizhu Chen},
  title        = {LoRA: Low-Rank Adaptation of Large Language Models},
  booktitle    = {The Tenth International Conference on Learning Representations, {ICLR}
                  2022, Virtual Event, April 25-29, 2022},
  publisher    = {OpenReview.net},
  year         = {2022},
  url          = {https://openreview.net/forum?id=nZeVKeeFYf9},
}

@misc{liu2024llavanext,
    title={LLaVA-NeXT: Improved reasoning, OCR, and world knowledge},
    url={https://llava-vl.github.io/blog/2024-01-30-llava-next/},
    author={Liu, Haotian and Li, Chunyuan and Li, Yuheng and Li, Bo and Zhang, Yuanhan and Shen, Sheng and Lee, Yong Jae},
    month={January},
    year={2024}
}

@inproceedings{DBLP:conf/nips/DettmersPHZ23,
  author       = {Tim Dettmers and
                  Artidoro Pagnoni and
                  Ari Holtzman and
                  Luke Zettlemoyer},
  title        = {QLoRA: Efficient Finetuning of Quantized LLMs},
  booktitle    = {Advances in Neural Information Processing Systems 36: Annual Conference
                  on Neural Information Processing Systems 2023, NeurIPS 2023, New Orleans,
                  LA, USA, December 10 - 16, 2023},
  year         = {2023}
}

@inproceedings{DBLP:conf/iclr/LoshchilovH19,
  author       = {Ilya Loshchilov and
                  Frank Hutter},
  title        = {Decoupled Weight Decay Regularization},
  booktitle    = {7th International Conference on Learning Representations, {ICLR} 2019,
                  New Orleans, LA, USA, May 6-9, 2019},
  publisher    = {OpenReview.net},
  year         = {2019},
  url          = {https://openreview.net/forum?id=Bkg6RiCqY7},
}

@inproceedings{lin2004rouge,
  title={Rouge: A package for automatic evaluation of summaries},
  author={Lin, Chin-Yew},
  booktitle={Text summarization branches out},
  pages={74--81},
  year={2004}
}

@article{DBLP:journals/corr/abs-2412-18416,
  author       = {Zihan Wang and
                  Xiaocui Yang and
                  Yongkang Liu and
                  Shi Feng and
                  Daling Wang and
                  Yifei Zhang},
  title        = {Muse: {A} Multimodal Conversational Recommendation Dataset with Scenario-Grounded
                  User Profiles},
  journal      = {CoRR},
  volume       = {abs/2412.18416},
  year         = {2024},
  url          = {https://doi.org/10.48550/arXiv.2412.18416},
  doi          = {10.48550/ARXIV.2412.18416},
}

@inproceedings{DBLP:conf/nips/ZhangYSJC19,
  author       = {Ruiyi Zhang and
                  Tong Yu and
                  Yilin Shen and
                  Hongxia Jin and
                  Changyou Chen},
  editor       = {Hanna M. Wallach and
                  Hugo Larochelle and
                  Alina Beygelzimer and
                  Florence d'Alch{\'{e}}{-}Buc and
                  Emily B. Fox and
                  Roman Garnett},
  title        = {Text-Based Interactive Recommendation via Constraint-Augmented Reinforcement
                  Learning},
  booktitle    = {Advances in Neural Information Processing Systems 32: Annual Conference
                  on Neural Information Processing Systems 2019, NeurIPS 2019, December
                  8-14, 2019, Vancouver, BC, Canada},
  pages        = {15188--15198},
  year         = {2019},
}

@inproceedings{DBLP:conf/acl/LuBSMCWH21,
  author       = {Yu Lu and
                  Junwei Bao and
                  Yan Song and
                  Zichen Ma and
                  Shuguang Cui and
                  Youzheng Wu and
                  Xiaodong He},
  title        = {RevCore: Review-Augmented Conversational Recommendation},
  booktitle    = {Findings of the Association for Computational Linguistics: {ACL/IJCNLP}
                  2021, Online Event, August 1-6, 2021},
  series       = {Findings of {ACL}},
  volume       = {{ACL/IJCNLP} 2021},
  pages        = {1161--1173},
  publisher    = {Association for Computational Linguistics},
  year         = {2021},
}

@inproceedings{DBLP:conf/sigir/DengL0DL21,
  author       = {Yang Deng and
                  Yaliang Li and
                  Fei Sun and
                  Bolin Ding and
                  Wai Lam},
  title        = {Unified Conversational Recommendation Policy Learning via Graph-based
                  Reinforcement Learning},
  booktitle    = {{SIGIR} '21: The 44th International {ACM} {SIGIR} Conference on Research
                  and Development in Information Retrieval, Virtual Event, Canada, July
                  11-15, 2021},
  pages        = {1431--1441},
  publisher    = {{ACM}},
  year         = {2021},
  url          = {https://doi.org/10.1145/3404835.3462913},
  doi          = {10.1145/3404835.3462913},
}

@article{DBLP:journals/corr/abs-2110-07477,
  author       = {Lingzhi Wang and
                  Huang Hu and
                  Lei Sha and
                  Can Xu and
                  Kam{-}Fai Wong and
                  Daxin Jiang},
  title        = {Finetuning Large-Scale Pre-trained Language Models for Conversational
                  Recommendation with Knowledge Graph},
  journal      = {CoRR},
  volume       = {abs/2110.07477},
  year         = {2021},
  url          = {https://arxiv.org/abs/2110.07477},
}

@article{DBLP:journals/corr/abs-2408-12574,
  author       = {Haojun Shi and
                  Suyu Ye and
                  Xinyu Fang and
                  Chuanyang Jin and
                  Leyla Isik and
                  Yen{-}Ling Kuo and
                  Tianmin Shu},
  title        = {MuMA-ToM: Multi-modal Multi-Agent Theory of Mind},
  journal      = {CoRR},
  volume       = {abs/2408.12574},
  year         = {2024},
  url          = {https://doi.org/10.48550/arXiv.2408.12574},
  doi          = {10.48550/ARXIV.2408.12574},
}

@inproceedings{holtzman2020curious,
  title={The Curious Case of Neural Text Degeneration},
  author={Holtzman, Ari and Buys, Jan and Du, Li and Forbes, Maxwell and Choi, Yejin},
  booktitle={International Conference on Learning Representations},
  year={2020}
}

@inproceedings{DBLP:conf/recsys/NieZYDZCZCLCXH24,
  author       = {Guangtao Nie and
                  Rong Zhi and
                  Xiaofan Yan and
                  Yufan Du and
                  Xiangyang Zhang and
                  Jianwei Chen and
                  Mi Zhou and
                  Hongshen Chen and
                  Tianhao Li and
                  Ziguang Cheng and
                  Sulong Xu and
                  Jinghe Hu},
  title        = {A Hybrid Multi-Agent Conversational Recommender System with {LLM}
                  and Search Engine in E-commerce},
  booktitle    = {Proceedings of the 18th {ACM} Conference on Recommender Systems, RecSys
                  2024, Bari, Italy, October 14-18, 2024},
  pages        = {745--747},
  publisher    = {{ACM}},
  year         = {2024},
  url          = {https://doi.org/10.1145/3640457.3688061},
  doi          = {10.1145/3640457.3688061},
}

@inproceedings{DBLP:conf/ecir/HouZLLXMZ24,
  author       = {Yupeng Hou and
                  Junjie Zhang and
                  Zihan Lin and
                  Hongyu Lu and
                  Ruobing Xie and
                  Julian J. McAuley and
                  Wayne Xin Zhao},
  title        = {Large Language Models are Zero-Shot Rankers for Recommender Systems},
  booktitle    = {Advances in Information Retrieval - 46th European Conference on Information
                  Retrieval, {ECIR} 2024, Glasgow, UK, March 24-28, 2024, Proceedings,
                  Part {II}},
  series       = {Lecture Notes in Computer Science},
  volume       = {14609},
  pages        = {364--381},
  publisher    = {Springer},
  year         = {2024},
  url          = {https://doi.org/10.1007/978-3-031-56060-6\_24},
}

@misc{qwen2.5-VL,
    title = {Qwen2.5-VL},
    url = {https://qwenlm.github.io/blog/qwen2.5-vl/},
    author = {Qwen Team},
    month = {January},
    year = {2025}
}

@article{yang2025reges,
  title         = {ReGeS: Reciprocal Retrieval-Generation Synergy for Conversational Recommender Systems},
  author        = {Yang, Dayu and Fang, Hui},
  year          = {2025},
  month         = {sep},
  journal       = {arXiv preprint arXiv:2509.21371},
  archivePrefix = {arXiv},
  eprint        = {2509.21371},
  primaryClass  = {cs.IR},
  doi           = {10.48550/arXiv.2509.21371},
  url           = {https://arxiv.org/abs/2509.21371},
  note          = {Accepted at WISE 2025}
}

@article{baker2009action,
  title={Action understanding as inverse planning},
  author={Baker, Chris L and Saxe, Rebecca and Tenenbaum, Joshua B},
  journal={Cognition},
  volume={113},
  number={3},
  pages={329--349},
  year={2009},
  publisher={Elsevier}
}

@inproceedings{ullman2009help,
  title={Help or hinder: Bayesian models of social goal inference},
  author={Ullman, Tomer and Baker, Chris and Macindoe, Owen and Evans, Owain and Goodman, Noah and Tenenbaum, Joshua},
  booktitle={Advances in neural information processing systems},
  volume={22},
  year={2009}
}

@inproceedings{wang-etal-2024-instruct,
    title = "Instruct Once, Chat Consistently in Multiple Rounds: An Efficient Tuning Framework for Dialogue",
    author = "Wang, Jian  and
      Leong, Chak Tou  and
      Wang, Jiashuo  and
      Lin, Dongding  and
      Li, Wenjie  and
      Wei, Xiaoyong",
    editor = "Ku, Lun-Wei  and
      Martins, Andre  and
      Srikumar, Vivek",
    booktitle = "Proceedings of the 62nd Annual Meeting of the Association for Computational Linguistics (Volume 1: Long Papers)",
    month = aug,
    year = "2024",
    address = "Bangkok, Thailand",
    publisher = "Association for Computational Linguistics",
    url = "https://aclanthology.org/2024.acl-long.219/",
    doi = "10.18653/v1/2024.acl-long.219",
    pages = "3993--4010"
}

@inproceedings{hausknecht2015deep,
  title={Deep Recurrent Q-Learning for Partially Observable MDPs.},
  author={Hausknecht, Matthew J and Stone, Peter},
  booktitle={AAAI fall symposia},
  volume={45},
  pages={141},
  year={2015}
}

@inproceedings{rabinowitz2018machine,
  title={Machine theory of mind},
  author={Rabinowitz, Neil and Perbet, Frank and Song, Francis and Zhang, Chiyuan and Eslami, SM Ali and Botvinick, Matthew},
  booktitle={International conference on machine learning},
  pages={4218--4227},
  year={2018},
  organization={PMLR}
}

@inproceedings{DBLP:conf/emnlp/KarpukhinOMLWEC20,
  author       = {Vladimir Karpukhin and
                  Barlas Oguz and
                  Sewon Min and
                  Patrick Lewis and
                  Ledell Wu and
                  Sergey Edunov and
                  Danqi Chen and
                  Wen{-}tau Yih},
  editor       = {Bonnie Webber and
                  Trevor Cohn and
                  Yulan He and
                  Yang Liu},
  title        = {Dense Passage Retrieval for Open-Domain Question Answering},
  booktitle    = {Proceedings of the 2020 Conference on Empirical Methods in Natural
                  Language Processing, {EMNLP} 2020, Online, November 16-20, 2020},
  pages        = {6769--6781},
  publisher    = {Association for Computational Linguistics},
  year         = {2020},
  url          = {https://doi.org/10.18653/v1/2020.emnlp-main.550},
  doi          = {10.18653/V1/2020.EMNLP-MAIN.550},
  bibsource    = {dblp computer science bibliography, https://dblp.org}
}

\newpage
\appendix
\section{Datasets and Preprocessing}
\label{appendix:datasets}

We evaluate our method using two publicly available SCR datasets: \textbf{SIMMC 2.1}~\citep{kottur2023overview} and \textbf{SCREEN}~\citep{DBLP:conf/mm/LinWLL24}. 
The SIMMC 2.1 dataset provides a multimodal, task-oriented dialogue corpus that captures interactions between customers and sales assistants within an immersive 3D virtual shopping environment. The SCREEN dataset comprises over 20,000 synthetic dialogues focused on situated conversational recommendations.
To better align with the real-world SCR setting and enable a more holistic evaluation, we construct a balanced test set by sampling dialogues with and without scene transitions in a 1:1 ratio. This ensures that both static and dynamic situational contexts are well covered for evaluation. Table \ref{tab:dataset:statistics} presents detailed statistics for the two datasets.

\begin{table}[th!]
\centering
\resizebox{0.99\linewidth}{!}{
\begin{tabular}{lcc}
\toprule
Dataset & SIMMC 2.1 & SCREEN \\
\midrule
Total \#dialogue & 5,622 & 20,081 \\
Total \#utterances & 58,717 & 190,011 \\
Total \#scene snapshots & 1,566 & 1,566 \\
Avg.  \#words per user turns & 12.6 & 22.46 \\
Avg.  \#words per system turns & 13.4 & 33.41 \\
Avg.  \#utterances per dialog & 10.4 & 9.46 \\
Avg.  \#objects mentioned per dialog & 4.7 & 4.4 \\
Avg.  \#objects in scene & 19.7 & 19.7 \\
\bottomrule
\end{tabular}
}
\caption{Statistics of the experimental datasets.}
\label{tab:dataset:statistics}
\end{table}

Beyond the raw corpus size, we also quantify the two central challenges studied in this paper. First, dynamic scene transitions are a frequent phenomenon in our evaluation setting: by construction, 50\% of the dialogues in the repurposed test split require moving to a different scene, which ensures that transition reasoning is not a marginal corner case. Second, implicit preference discovery is particularly prominent in SCREEN. We identify dialogues whose initial user request does not fully specify the target item attributes and therefore requires refinement through later interaction. Under this definition, over 90\% of SCREEN dialogues require the system to progressively infer implicit preferences from multi-turn conversational feedback.

\section{Baseline Implementation Details}
\label{appendix:model_details}

We provide the specific implementation configurations for the baseline models, categorized by their learning paradigms. We also clarify the exact input modalities made available to each model family for fair comparison.

\paragraph{Inference-only Baselines (CoT \& ICL).}
We evaluate three backbone MLLMs in inference-only modes: GPT-4o~\citep{openai2024gpt4o}, LLaVA-NeXT-7B~\citep{liu2024llavanext}, and Qwen2.5-VL-7B-Instruct~\citep{DBLP:journals/corr/abs-2409-12191}.
\begin{itemize}[leftmargin=*]
    \item \textbf{Settings:} For the open-source models (LLaVA-NeXT and Qwen2.5-VL), we utilize 4-bit quantization to optimize memory usage. Inference is conducted with a temperature of 0.7 and a maximum token limit of 256. All inference-only vision-language baselines receive the raw scene image together with the dialogue history and the textual item metadata of the current scene.
    \item \textbf{In-Context Learning (ICL):} For ICL settings, we retrieve the top-$k$ ($k=2$) semantically similar demonstrations using a dense encoder (consistent with our STE module) and prepend them to the conversation history as guidance.
\end{itemize}

\paragraph{Training-based Baselines.}
We compare our method against fully supervised baselines, including small-scale multimodal models, text-based recommenders, and fine-tuned MLLMs.
\begin{itemize}[leftmargin=*]
    \item \textbf{Small Multimodal Models (ALBEF):} We utilize ALBEF~\citep{DBLP:conf/nips/LiSGJXH21} as a representative small-scale baseline. It is optimized via \textit{full-parameter fine-tuning} using a sequence-to-sequence objective. As a vision-language baseline, it is given access to the same scene image and dialogue context as other multimodal methods. We set the learning rate to $1 \times 10^{-5}$, batch size to 32, and train for 10 epochs.
    
    \item \textbf{Text-based Recommender (ReGeS):} For ReGeS~\citep{yang2025reges}, since it cannot process visual inputs, we convert visual scenes into structured textual profiles using captions generated by GPT-4o. This conversion is performed once offline and reused during both training and evaluation so that ReGeS has access to the same environment information in textual form. We fine-tune ReGeS using LoRA~\citep{DBLP:conf/iclr/HuSWALWWC22} for efficient adaptation.
    
    \item \textbf{Fine-tuned MLLMs:} We also report the performance of LLaVA-NeXT and Qwen2.5-VL under supervised fine-tuning. These models use the same multimodal inputs as their inference-only counterparts. Similar to ReGeS, these large models are optimized using LoRA rather than full-parameter tuning to maintain computational efficiency. The LoRA rank is set to $r=64$ with scaling $\alpha=16$.
\end{itemize}
All training-based baselines are tuned on the training set, and the best checkpoints are selected based on Recall@1 performance on the validation set.

\section{Additional Implementation Details}
\label{appendix:implementation_details}
We provide further details on the model configuration and training process to ensure reproducibility. The only proprietary component in our pipeline, GPT-4o, is used in a one-time offline preprocessing step to build textual scene profiles. All online transition estimation, preference inference, and response generation rely on the open-source components listed below.
\paragraph{Scene Transition Estimation (STE).}
For the STE module, we utilize Qwen3-Embedding-4B as the dense retriever and Qwen3-Reranker-4B as the fine-grained reranker. The generated target profile is never executed directly as a transition; it is used only as a semantic query against the candidate scene pool. To optimize the reranker, we apply LoRA fine-tuning to the attention modules (query and value projections). The LoRA rank is set to $r=64$ with a scaling factor $\alpha=16$ and a dropout rate of $0.05$. We train the reranker for 3 epochs with a batch size of 8.
\paragraph{Bayesian Inverse Inference (BI-INF).}
For the preference reasoning module, we fine-tune the Qwen2.5-VL-7B-Instruct backbone. To mitigate memory constraints during training on NVIDIA A100 GPUs, we employ 4-bit quantization via BitsAndBytes \citep{DBLP:conf/nips/DettmersPHZ23}. The LoRA configuration aligns with the STE module ($r=64, \alpha=16$), targeting all linear layers in the vision-language projection and attention mechanisms. We set the maximum input length to 3,000 tokens to accommodate multi-turn dialogue history and visual features, while the output length is restricted to 256 tokens. The model is optimized for 3 epochs with a batch size of 16 (using gradient accumulation).
\paragraph{Inference and Generation.}
During the inference phase, BI-INF scores candidate items by computing the likelihood of the observed dialogue state under competing hypotheses from the model logits, and response generation is invoked only after the candidates are ranked. We then employ nucleus sampling \citep{holtzman2020curious} to generate diverse and natural responses. We set the cumulative probability threshold top-$p$ to 0.75 and the candidate pool size top-$k$ to 40. The maximum decoding length is set to 256 tokens. All detailed hyperparameters are summarized in Table \ref{tab:parameters}.

\begin{table}[t!]
\centering
\resizebox{0.96\linewidth}{!}{
\begin{tabular}{l c}
\toprule
\textbf{Parameter} & \textbf{Value} \\
\midrule
\multicolumn{2}{c}{\textit{Model Architectures}} \\
\midrule
Backbone MLLM (BI-INF) & Qwen2.5-VL-7B-Instruct \\
STE Dense Encoder & Qwen3-Embedding-4B \\
STE Reranker & Qwen3-Reranker-4B \\
Scene Captioning & GPT-4o \\
\midrule
\multicolumn{2}{c}{\textit{Training Optimization}} \\
\midrule
LoRA Rank $r$ & 64 \\
LoRA Scaling $\alpha$ & 16 \\
LoRA Dropout & 0.05 \\
Quantization & 4-bit (BitsAndBytes) \\
Optimizer & AdamW \\
Learning Rate & $2 \times 10^{-5}$ \\
Warm-up Ratio & 0.03 \\
Weight Decay & 0.01 \\
Epochs & 3 \\
Batch Size (STE) & 8 \\
Batch Size (BI-INF) & 16 \\
\midrule
\multicolumn{2}{c}{\textit{Context Lengths}} \\
\midrule
Max. tokens for STE input & 512 \\
Max. tokens for BI-INF input & 3000 \\
Max. tokens for BI-INF output & 256 \\
\midrule
\multicolumn{2}{c}{\textit{Inference \& Generation}} \\
\midrule
Decoding Strategy & Nucleus Sampling \\
Top-$p$ & 0.75 \\
Top-$k$ & 40 \\
Temperature & 0.7 \\
Max. tokens for response & 256 \\
\bottomrule
\end{tabular}}
\caption{Detailed hyperparameters and experimental settings for training and inference.}
\label{tab:parameters}
\end{table}

\section{Additional Efficiency and Transition Analyses}
\label{appendix:additional_analyses}

This section presents five supplementary analyses: end-to-end efficiency, a Qwen3 retrieval-only transition baseline, a global-access MLLM baseline, conditioned error propagation from STE to BI-INF, and a boundary-case discussion of STE. The latency breakdown and scene-density scaling results are reported in Tables~\ref{tab:efficiency_breakdown} and~\ref{tab:latency_bucket}, respectively. The retrieval-only comparison and the global-access baseline are summarized in Tables~\ref{tab:retrieval_only_baseline} and~\ref{tab:global_access_baseline}, respectively, the conditioned error-propagation analysis is reported in Table~\ref{tab:error_propagation_conditioned}, and the qualitative boundary-case discussion follows afterward.

\paragraph{Latency Breakdown and Scene-Density Scaling.}
The first analysis reports the component-wise latency of \model on SIMMC 2.1 and compares it with a strong trained MLLM baseline. The second analysis groups evaluation instances by the number of in-scene candidate items to show how BI-INF scales as scene density increases.

\begin{table}[t]
\centering
\small
\begin{tabular}{lcc}
\toprule
\textbf{Metric} & \makecell{\textbf{Qwen2.5-VL}\\\textbf{(Trained)}} & \makecell{\textbf{\model}\\\textbf{(Ours)}} \\
\midrule
STE Latency & N/A & 118 ms \\
BI-INF Latency & N/A & 245 ms \\
Generation Latency & 1427 ms & 1219 ms \\
Total Latency & 1427 ms & 1582 ms \\
R@1 & 29.47 & 38.75 \\
\bottomrule
\end{tabular}
\caption{Component-wise latency comparison between \model and a strong trained MLLM baseline.}
\label{tab:efficiency_breakdown}
\end{table}

\begin{table}[t]
\centering
\resizebox{\linewidth}{!}{
\begin{tabular}{lccccc}
\toprule
\textbf{\# Items in Scene} & 5--10 & 10--15 & 15--20 & 20--25 & $>$25 \\
\midrule
\textbf{Latency (s / turn)} & $\sim$0.8 & $\sim$1.2 & $\sim$1.7 & $\sim$2.3 & $\sim$2.9 \\
\bottomrule
\end{tabular}}
\caption{Latency scaling of \model with respect to scene density.}
\label{tab:latency_bucket}
\end{table}

\paragraph{Qwen3 Retrieval-only Transition Baseline.}
To separate the effect of backbone capacity from the effect of explicit target-profile generation, we compare \model with a retrieval-only transition strategy built on the identical Qwen3-Embedding-4B dense retrieval~\cite{DBLP:conf/emnlp/KarpukhinOMLWEC20} backbone used in STE. This baseline directly encodes the dialogue history and retrieves the target scene from the global pool, but removes the intermediate target-profile generation step and the subsequent generative reasoning. As summarized in Table~\ref{tab:retrieval_only_baseline}, a direct MLLM decision over all candidate scenes is computationally infeasible when the environment contains 1,566 scene snapshots, so we report it as an infeasible reference point rather than a runnable quantitative baseline.

\begin{table}[t]
\centering
\resizebox{\linewidth}{!}{
\begin{tabular}{lcc}
\toprule
\textbf{Transition Strategy} & \textbf{Feasibility} & \makecell{\textbf{Final Rec.}\\\textbf{R@1}} \\
\midrule
Direct MLLM Decision & Infeasible & N/A \\
Qwen3 Dense Retrieval & Feasible & 35.24 \\
\model STE & Feasible & 38.75 \\
\bottomrule
\end{tabular}}
\caption{Comparison of scene transition strategies on SIMMC 2.1. Direct MLLM decision over all 1,566 candidate scenes is included as an infeasible reference point, while Qwen3 dense retrieval and our generative-retrieval STE are both runnable strategies.}
\label{tab:retrieval_only_baseline}
\end{table}

\paragraph{Global-access MLLM Baseline.}
A potential concern is that the baseline MLLMs in Table~\ref{table:recommendation_result} only observe the current scene, whereas \model can search over the global scene pool through STE. To examine this setting more directly, we augment the strong zero-shot Qwen2.5-VL baseline with the top-5 scene images returned by our coarse retrieval stage and ask the model to jointly reason over these candidate scenes. We use the zero-shot version here to isolate the effect of expanded scene access alone, without conflating it with additional supervised adaptation. As shown in Table~\ref{tab:global_access_baseline}, this global-access variant improves R@1 from 16.72 to 19.58 on SIMMC 2.1, but it still remains far below the 38.75 achieved by \model. Moreover, packing five retrieved scenes into a single prompt substantially increases inference latency (+138.2\%), indicating that naively expanding the visual context is both less effective and less efficient than our decoupled STE+BI-INF design.

\begin{table}[t]
\centering
\small
\begin{tabular}{lcc}
\toprule
\textbf{Method} & \makecell{\textbf{Final Rec.}\\\textbf{R@1}} & \makecell{\textbf{Latency}\\\textbf{vs. Zero-shot}} \\
\midrule
\makecell[l]{Qwen2.5-VL\\(current scene)} & 16.72 & \makecell{Base\\(1427 ms)} \\
\makecell[l]{Qwen2.5-VL\\(+ top-5 scenes)} & 19.58 & +138.2\% \\
\makecell[l]{\model\\(single filtered scene)} & 38.75 & +10.8\% \\
\bottomrule
\end{tabular}
\caption{Comparison with a global-access Qwen2.5-VL baseline on SIMMC 2.1.}
\label{tab:global_access_baseline}
\end{table}

\paragraph{Conditioned Error Propagation from STE to BI-INF.}
The \textit{w/ Non-target} ablation in the main paper already shows that feeding BI-INF an incorrect scene substantially degrades recommendation quality. To make this dependency more explicit, we further condition the evaluation on whether STE predicts the correct target scene on SIMMC 2.1, thereby separating errors caused by scene grounding from errors caused by in-scene preference ranking. As shown in Table~\ref{tab:error_propagation_conditioned}, correct scene grounding yields 40.0 R@1 and 48.2 MRR@5, whereas these numbers fall to 29.8 and 40.1 once the scene prediction is incorrect. This 10.2-point drop in R@1 confirms that scene-estimation errors propagate directly to downstream item ranking, while the remaining gap between the conditioned and overall results also indicates that BI-INF still contributes non-trivial discrimination after the scene is correctly grounded.

\begin{table}[t]
\centering
\resizebox{0.85\linewidth}{!}{
\begin{tabular}{lcc}
\toprule
\textbf{Condition} & \textbf{R@1} & \textbf{MRR@5} \\
\midrule
Correct Scene Prediction & 40.0 & 48.2 \\
Incorrect Scene Prediction & 29.8 & 40.1 \\
\bottomrule
\end{tabular}}
\caption{Recommendation performance on SIMMC 2.1 conditioned on whether STE predicts the correct target scene.}
\label{tab:error_propagation_conditioned}
\end{table}

\paragraph{Boundary Cases of STE.}
Beyond the aggregate metrics above, we explicitly inspect the failure patterns of STE. We observe two recurring boundary cases. First, errors tend to arise when the user provides only sparse transition cues (e.g., a broad style request without distinctive attributes), in which case multiple candidate scenes remain semantically plausible after coarse retrieval. Second, mistakes also occur when visually similar scenes share overlapping inventories, causing the generated target profile to overemphasize high-level semantics while under-specifying the decisive fine-grained attributes. Importantly, these failures do not lead to unconstrained hallucinated transitions: the coarse-to-fine design still grounds the final prediction in a real scene from the candidate pool, and the downstream recommendation quality degrades gracefully rather than collapsing. This pattern is consistent with the \textit{w/ Non-target} ablation in Table~\ref{tab:robustness_transitions}, the conditioned analysis in Table~\ref{tab:error_propagation_conditioned}, and the qualitative example in Figure~\ref{fig:case_study2}, where baselines without reliable transition reasoning are more likely to end in passive or erroneous responses. We therefore view the main remaining challenge of STE not as free-form hallucination, but as disambiguating among semantically neighboring real scenes under limited conversational evidence.

\section{Prompting Templates}
\label{appendix:prompt}

We present the detailed templates used for data construction and model training. First, to ensure standard dialogue state tracking, we adhere to a predefined schema of intents and slots. As illustrated in Figure \ref{fig:predefined_intent_slot}, representative intents include requesting product details (i.e., \texttt{REQUEST:GET}), comparing items (i.e., \texttt{REQUEST:COMPARE}), and adding items to the shopping cart (i.e., \texttt{REQUEST:ADD\_TO\_CART}). Corresponding slots may include attributes such as customer review, color, and price.

Furthermore, regarding the Bayesian Inverse Inference module, we structure the instruction tuning data as shown in Figure \ref{fig:MLLM_Finetune_Template}. This template integrates the visual scene snapshots, the dialogue history, and hypothetical user preferences (like/dislike) to formulate the input, while the output is the corresponding dialogue state. This structured format enables the MLLM to learn the inverse mapping from user goals to dialogue actions effectively.

\section{Prompt for GPT-Score Evaluation} 
\label{appendix:gpt_score}

To ensure a comprehensive evaluation of the generated responses, we employ GPT-4o as an unbiased judge. The evaluation prompt is designed to assess the response quality based on three critical dimensions: \textbf{Relevance} (whether the response addresses the user's intent), \textbf{Visual Grounding} (whether the mentioned items and attributes align with the provided scene), and \textbf{Fluency} (whether the text is natural and coherent). The full prompt template is presented in Figure~\ref{fig:gpt_eval_prompt}.

\begin{figure}[t!]
\centering
\small
\begin{tcolorbox}[
    colback=gray!10!white,
    colframe=black!75!black,
    title=Predefined Schema of Intents and Slots,
    fonttitle=\bfseries
]
\textbf{Intents:}

(1) INFORM: GET, \\
(2) REQUEST: COMPARE, \\
(3) REQUEST: ADD\_TO\_CART, \\
(4) INFORM: REFINE, \\
(5) REQUEST: DISAMBIGUATE, \\
(6) ASK: GET, \\
(7) INFORM: DISAMBIGUATE, \\
(8) REQUEST: GET ...

\vspace{1em}
\textbf{Slots:}

(1) ASSET TYPE, (2) CUSTOMER REVIEW, (3) AVAILABLE SIZES, (4) COLOR, 
(5) PATTERN, (6) BRAND, (7) SLEEVE LENGTH, (8) TYPE, (9) PRICE, 
(10) SIZE, (11) CUSTOMER RATING, (12) MATERIALS ...

\end{tcolorbox}
\caption{Representative instances of the predefined intents and slots across SCR datasets.}
\label{fig:predefined_intent_slot}
\end{figure}

\begin{figure}[t!]
\centering
\small
\begin{tcolorbox}[
    colback=gray!10!white,
    colframe=black!75!black,
    title=Input-Output Format for Fine-tuning,
    fonttitle=\bfseries
]
\textbf{Input Instruction:}

The provided image consists of a series of scene snapshots, accompanied by hypothetical user preferences: \texttt{\{User like/dislike target item\}}. The items mentioned in the conversation history are as follows: \texttt{\{item1, item2, ...\}}. The information of the items in the scene: \texttt{\{item1 (attributes1), item2 (attributes2), ...\}}. Given the dialogue state of the conversation up to $n$-1 turns represented as: \texttt{\{\{Intent\_1, Slot\_1, Value\_1\}, ...\}}, the current dialogue state at the $n$-th turn is:

\vspace{1em}
\textbf{Output:}

\texttt{\{Intent\_n, Slot\_n, Value\_n\}}

\end{tcolorbox}
\caption{Input-Output format for fine-tuning the policy model in Bayesian inverse inference.}
\label{fig:MLLM_Finetune_Template}
\end{figure}

\begin{figure*}[ht] \centering \small \begin{tcolorbox}[colback=gray!10!white,colframe=black!75!black,title=Prompt Template for Evaluation] 
\textbf{System Instruction:} You are an expert judge for Situated Conversational Recommendation (SCR) systems. Your task is to evaluate the quality of a response generated by an AI assistant based on a user's request and a specific visual environment.
\\

\textbf{Input Context:} \begin{itemize}[leftmargin=*] \item \textbf{Visual Scene Description:} A structured text describing the items visible in the current environment (e.g., item IDs, types, colors, positions). \item \textbf{Dialogue History:} The conversation logs between the user and the assistant leading up to the current turn. \item \textbf{Ground Truth Response:} The human-annotated standard response. \item \textbf{Candidate Response:} The response generated by the model to be evaluated. \end{itemize}

\textbf{Input Data:} \begin{itemize}[leftmargin=*] \item \textbf{[SCENE]:} {SCENE\_PROFILE} \item \textbf{[HISTORY]:} {DIALOGUE\_HISTORY} \item \textbf{[GROUND TRUTH]:} {REFERENCE\_RESPONSE} \item \textbf{[CANDIDATE]:} {GENERATED\_RESPONSE} \end{itemize}

\textbf{Evaluation Criteria:} Please rate the \textbf{[CANDIDATE]} response on a scale of 1 to 10 based on the following criteria: \begin{enumerate} \item \textbf{Visual Grounding:} Does the response accurately reflect the items in the [SCENE]? Does it avoid hallucinating items or attributes not present in the environment? \item \textbf{Relevance \& Intent:} Does the response correctly identify the user's intent (e.g., recommendation, transition, Q\&A)? Does it recommend items similar to the [GROUND TRUTH] or meet the user's constraints? \item \textbf{Fluency \& Coherence:} Is the response grammatically correct and contextually natural? \end{enumerate}

\textbf{Output Format:} Output a JSON object with two fields: \begin{itemize} \item ``reasoning'': A brief explanation of the judgment (max 50 words). 
\item ``score'': An integer score from 1 (worst) to 10 (perfect). \end{itemize} \end{tcolorbox} \caption{The prompt template used for GPT-Score evaluation. The placeholders {SCENE\_PROFILE}, {DIALOGUE\_HISTORY}, {REFERENCE\_RESPONSE}, and {GENERATED\_RESPONSE} are replaced with the actual test data during evaluation.} \label{fig:gpt_eval_prompt} 
\end{figure*}

\section{Human Evaluation Details}
\label{appendix:human_evaluation}

We conducted a human evaluation to assess the quality of generated responses. Specifically, we randomly selected 30 test instances from the \textbf{SCREEN} test set, then recruited three well-educated graduate students to serve as annotators. For each instance, the annotators rated the responses produced by different models on a 3-point ordinal scale (0–2), where 0=\emph{Weak}, 1=\emph{Moderate}, and 2=\emph{Excellent}. Ratings were given along three dimensions:
\begin{enumerate}[label=(\arabic*), leftmargin=*, labelindent=0pt, labelsep=0.5em, align=left]
  \item \textbf{Coherence (Coher.)}: logical flow and internal consistency of the response.
  \begin{itemize}[left=0pt]
    \item \emph{Weak (0)}: disjoint or abrupt topic shifts; unclear links.
    \item \emph{Moderate (1)}: mostly logical with minor inconsistencies or awkward transitions.
    \item \emph{Excellent (2)}: clear progression and well-connected ideas throughout.
  \end{itemize}

  \item \textbf{Informativeness (Inform.)}: completeness and relevance of content to the user’s query and dialogue context.
  \begin{itemize}[left=0pt]
    \item \emph{Weak (0)}: minimal or vague information; key details missing.
    \item \emph{Moderate (1)}: some relevant details but incomplete coverage or specificity.
    \item \emph{Excellent (2)}: accurate, sufficient details that fully address the query/context.
  \end{itemize}

  \item \textbf{Situatedness (Situat.)}: degree of tailoring to the current dialogue state and scene; effective use of situational cues.
  \begin{itemize}[left=0pt]
    \item \emph{Weak (0)}: generic; ignores the specific context.
    \item \emph{Moderate (1)}: partial awareness of context with limited adaptation.
    \item \emph{Excellent (2)}: strongly context-aware and responsive to the user’s immediate needs and environment.
  \end{itemize}
\end{enumerate}

To quantify inter-annotator agreement, we adopt Fleiss’ kappa~\cite{fleiss1971measuring} computed per dimension.
Figure~\ref{fig:human_eval_radar} presents the human evaluation results. We observe that each obtained Fleiss' kappa falls into the range $[0.47, 0.53]$, indicating moderate agreement among annotators. Notably, our \model achieves the highest scores across all three metrics. 
These gains indicate that the integration of scene transition estimation and Bayesian inverse reasoning with ToM enables more faithful preference modeling. Consequently, it results in more coherent, informative responses that are contextually grounded.

\section{Case Study}
\label{appendix:case_study}

\begin{figure*}[htbp]
    \centering
    \includegraphics[width=0.95\linewidth]{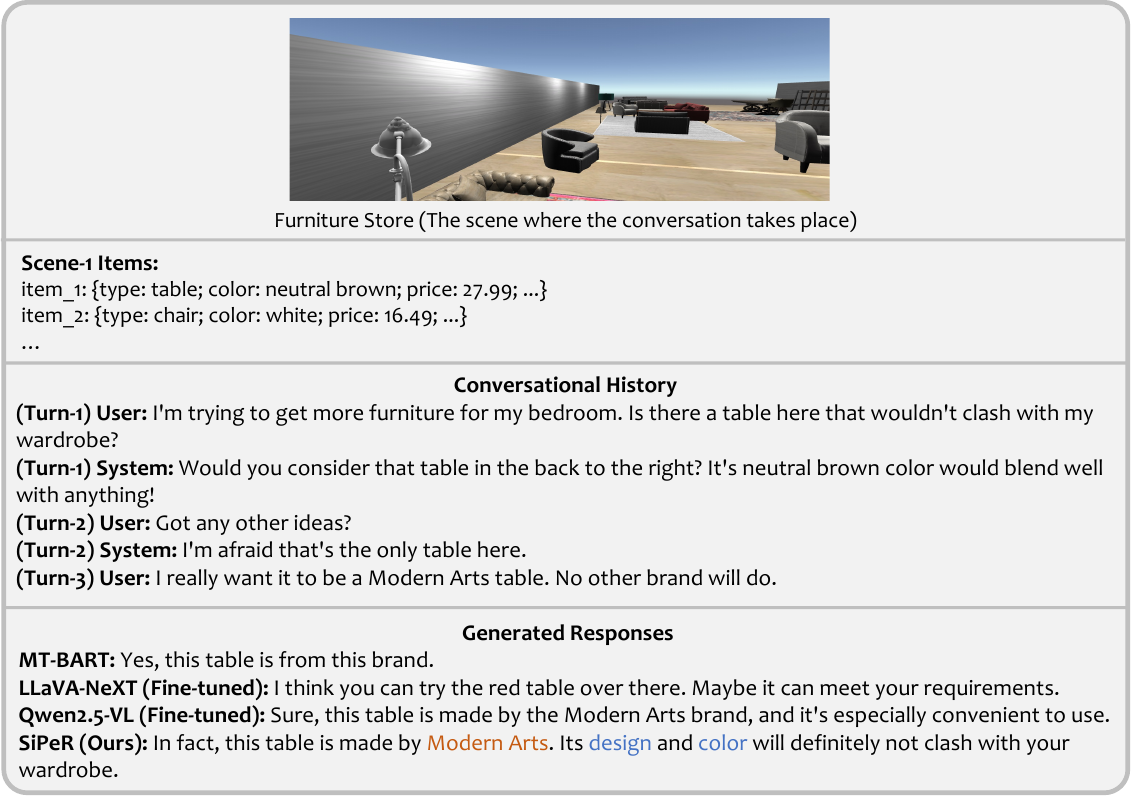}
    \caption{Cases of generated responses for different models, where no scene transition is needed.}
    \label{fig:case_study1}
\end{figure*}

\paragraph{Case A: No scene transition required (Figure~\ref{fig:case_study1}).} 
In this scenario, the user holds both an explicit constraint (brand \emph{Modern Arts}) and an implicit visual preference (compatibility with their \emph{wardrobe}). As shown in Figure~\ref{fig:case_study1}, baseline models exhibit distinct failure modes. The small-scale model \textbf{ALBEF} generates a generic caption, ignoring the specific brand constraint. The text-only baseline \textbf{ReGeS} suffers from hallucination, inventing a ``black'' table despite the system previously stating only a brown one exists. While the strong MLLM \textbf{Qwen2.5-VL} correctly identifies the brand, its response is factually rigid and fails to address the user's underlying concern about style compatibility. In contrast, \textbf{\model} successfully grounds the ``Modern Arts'' brand in the visual object and, crucially, links its neutral attributes back to the user's initial latent goal (``won't clash with my wardrobe''). This demonstrates the effectiveness of our Bayesian Inverse Inference in disentangling implicit intents from surface-level dialogue.

\begin{figure*}[hbpt]
    \centering
    \includegraphics[width=0.95\linewidth]{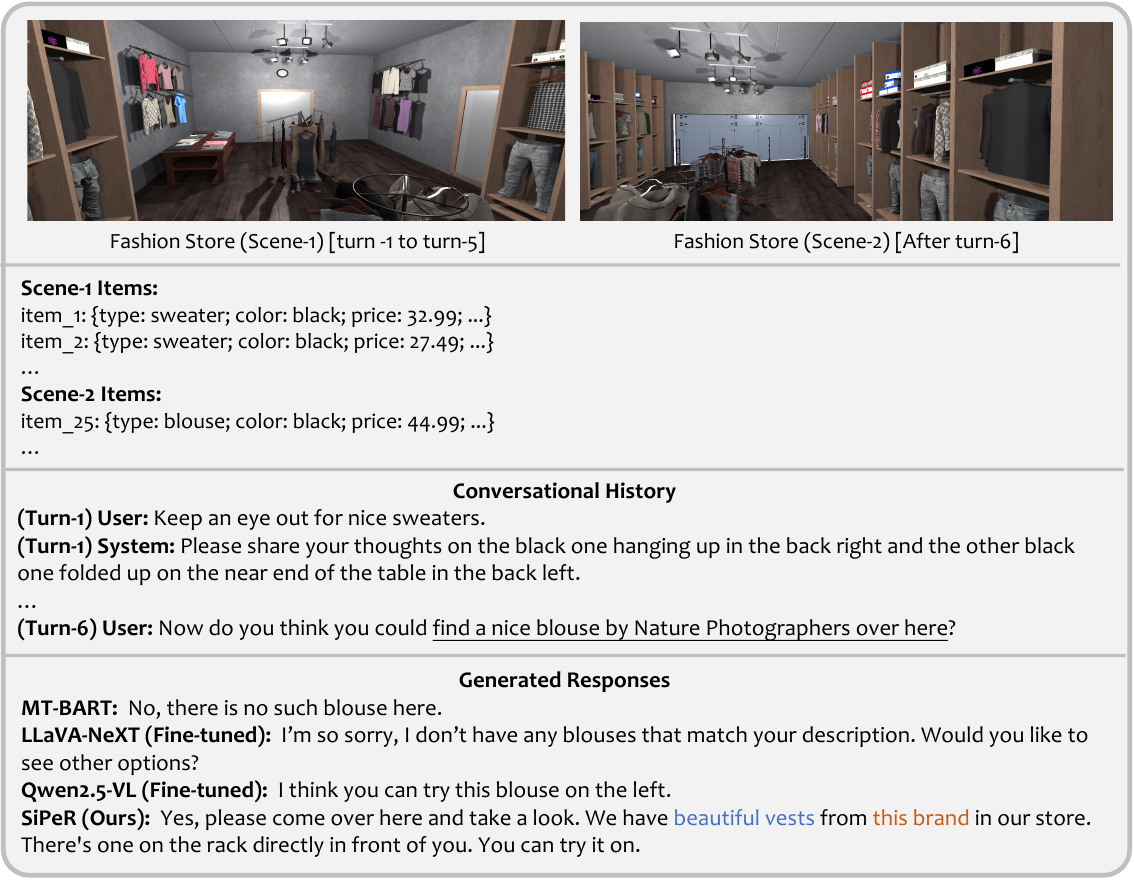}
    \caption{Cases of generated responses for different models when the scene transition is required.
    }
    \label{fig:case_study2}
\end{figure*}

\paragraph{Case B: Scene transition required (Figure~\ref{fig:case_study2}).} Figure~\ref{fig:case_study2} illustrates a dynamic scenario where the user requests a specific item (\emph{Nature Photographers} blouse) that is absent in the current location (\emph{Scene-1}, containing sweaters) but available in a different section (\emph{Scene-2}). Baselines lacking visual grounding or spatial awareness fail significantly: \textbf{ALBEF} remains anchored to the current view, merely describing the visible sweaters, while \textbf{ReGeS} hallucinates that the item is ``right here.'' Notably, even the advanced \textbf{Qwen2.5-VL}, while correctly recognizing the item's absence in the current scene, adopts a passive stance (``No blouses here''), resulting in a conversational dead-end. Conversely, \textbf{\model} leverages its Scene Transition Estimation (STE) mechanism to detect the mismatch between the user's request and the current environment. It proactively guides the user to the correct location (``come over here'') and accurately grounds the target item in the new scene, showcasing superior situatedness and navigational capability.

\end{document}